\documentclass[letterpaper, 10 pt, journal, twoside]{IEEEtran}
\usepackage{amsmath,amsfonts}
\usepackage{array}

\usepackage{enumerate}
\usepackage[utf8]{inputenc}
\usepackage[T1]{fontenc}
\usepackage{amsfonts}
\usepackage{amssymb}
\usepackage{tabularx}
\makeatletter
\let\NAT@parse\undefined
\makeatother
\usepackage{hyperref}
\hypersetup{
    colorlinks=true,
    linkcolor=blue,
    filecolor=magenta,      
    urlcolor=cyan,
    }
\usepackage{algorithm}  
\usepackage{algpseudocode}  
\usepackage{amsmath}  
\usepackage{breqn}
\usepackage[caption=false]{subfig}
\usepackage{textcomp}
\usepackage{stfloats}
\usepackage{url}
\usepackage{verbatim}
\usepackage{graphicx}
\usepackage{cite}
\usepackage{color}
\usepackage{colortbl}
\usepackage[most]{tcolorbox}
\let\labelindent\relax
\usepackage{enumitem}
\usepackage{multirow}
\usepackage{diagbox}
\usepackage{booktabs}
\usepackage{xcolor}
\usepackage{listings}
\definecolor{lightblue}{rgb}{0.68, 0.85, 0.90}
\definecolor{lightpurple}{rgb}{0.87, 0.81, 0.95}
\definecolor{lightlightgray}{rgb}{0.95,0.95,0.95}
\definecolor{lightpink}{rgb}{1.0, 0.87, 0.87}
\definecolor{darkgreen}{rgb}{0.0, 0.5, 0.0}
\definecolor{darkblue}{rgb}{0.0, 0.0, 0.5}
\definecolor{orange}{rgb}{1.0, 0.5, 0.0}
\definecolor{purple}{rgb}{0.5, 0.0, 0.5}
\definecolor{darkred}{rgb}{0.6, 0.0, 0.0}
\title{PrefCLM: Enhancing Preference-based Reinforcement Learning with \\Crowdsourced Large Language Models}

\author{Ruiqi Wang, Dezhong Zhao, Ziqin Yuan, Ike Obi, and Byung-Cheol Min
\thanks{Manuscript received: July, 10, 2024; Revised November, 24, 2024; Accepted January, 06 2025.}
\thanks{This paper was recommended for publication by Editor Gentiane Venture upon evaluation of the Associate Editor and Reviewers' comments.
This work involved human subjects in its research. The authors confirm that all human subject research procedures and protocols are exempt from review board approval. \textit{(Ruiqi Wang and Dezhong Zhao contributed equally to this work.)} \textit{(Corresponding author: Ruiqi Wang.)}} 
\thanks{Ruiqi Wang, Ziqin Yuan, Ike Obi, and Byung-Cheol Min are with the SMART Laboratory, Department of Computer and Information Technology, Purdue University, West Lafayette, IN, USA. {\tt\small{[wang5357, yuan460, obii, minb]@purdue.edu}.}}
\thanks{Dezhong Zhao is with the College of Mechanical and Electrical Engineering, Beijing University of Chemical Technology, Beijing, China. \textit{(This work was conducted during Dezhong Zhao was as a visiting scholar at the SMART Laboratory.)} \tt\small{DZ\_Zhao@buct.edu.cn}.}
\thanks{Digital Object Identifier (DOI): see top of this page.}
}

\begin{document}

\setlength{\abovedisplayskip}{1pt} 
\setlength{\belowdisplayskip}{1pt} 

\maketitle

\begin{abstract}
Preference-based reinforcement learning (PbRL) is emerging as a promising approach to teaching robots through human comparative feedback, sidestepping the need for complex reward engineering. \textcolor{black}{However, the substantial volume of human feedback required in existing PbRL methods hinders broader applications.} In this work, we introduce PrefCLM, a novel framework that utilizes crowdsourced large language models (LLMs) as synthetic teachers in PbRL. We utilize Dempster-Shafer Theory to fuse individual preferences from multiple LLM agents at the score level, efficiently leveraging their diversity and collective intelligence. We also introduce a human-in-the-loop pipeline, enabling iterative and collective refinements that adapt to the nuanced and individualized preferences inherent to human-robot interaction (HRI) scenarios. Experimental results across various general RL tasks show that PrefCLM achieves competitive performance compared to expert-engineered scripted teachers and excels in facilitating more more natural and efficient behaviors. A real-world user study (N=10) further demonstrates its capability to tailor robot behaviors to individual user preferences, significantly enhancing user satisfaction in HRI scenarios. Paper website: \url{https://prefclm.github.io}.
\end{abstract}


\section{Introduction}

Reinforcement learning (RL) has achieved notable success in robot learning; however, the effectiveness of learned policies relies heavily on well-designed reward functions \cite{zhu2019ingredients}. Crafting dense, informative rewards is challenging, requiring significant effort and domain expertise, particularly for long-horizon tasks \cite{wang2024initial} or tasks involving subtle social interactions \cite{wang2022feedback}. Additionally, reward hacking remains a potential risk \cite{hadfield2017inverse}.

In light of these challenges, preference-based reinforcement learning (PbRL) \cite{christiano2017deep,lee2021pebble,lee2021b,park2021surf,metcalf2023sample,liu2022task,liu2022meta,zhao2024prefmmt,kim2022preference} has emerged as a promising alternative by circumventing complex reward engineering. In PbRL, a human teacher provides comparative feedback on two distinct robot trajectories, each comprising multiple state-action pairs. A reward model is then learned to align with these preferences, guiding robot behaviors to better meet human expectations through RL. Despite recent advancements in enhancing feedback efficiency, the need for extensive human input remains a barrier to broader adoption.

\begin{figure}[t]
\centering
\includegraphics[width=0.9\columnwidth]{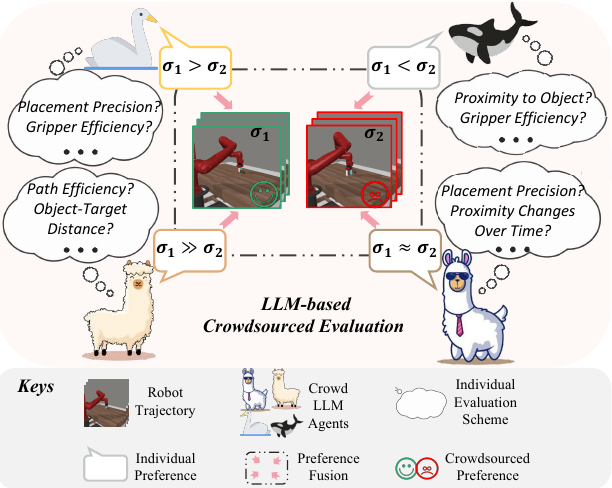}
\vspace{-8pt}
\caption{Conceptual illustration of the LLM-based crowdsourced evaluation. LLM instructors are symbolized by animal icons, with distinct species presenting varied LLM architectures. These crowd instructors, through their unique evaluative criteria and reasoning, determine individual preferences for robot trajectories, which are fused to formulate a unified crowdsourced preference used for PbRL process.}
\vspace{-15pt}
\label{fig:concept}
\end{figure}

\textcolor{black}{In response, synthetic feedback has been explored as a substitute for real human preferences in PbRL. A common approach employs scripted teachers \cite{lee2021b}, which generate synthetic preference labels based on cumulative reward values derived from predefined reward functions, to facilitate more efficient and rigorous PbRL algorithm benchmarking \cite{christiano2017deep,lee2021pebble,lee2021b,park2021surf,metcalf2023sample,liu2022task,liu2022meta}. While effective for controlled experimentation, the reintroduction of reward engineering and the static nature of scripted teachers limit their applicability in real-world settings, particularly in human-robot interaction (HRI) tasks where dynamic, individualized feedback is essential. }

\textcolor{black}{Another recent line of research explores the use of foundation agents, such as vision-language models (VLMs) and large language models (LLMs), as sources of synthetic feedback. For example, in \cite{wangrl}, a VLM agent is employed to analyze two image sequences simultaneously, each from a trajectory in comparison, and generate synthetic preference labels. This approach, while efficient for visually-driven tasks, may struggle to capture nuanced aspects of task execution, such as action smoothness, intent, and collaborative quality, which are critical in more complex tasks and HRI scenarios, due to its reliance solely on state images without action information. Additionally, some works have explored the use of a coding-LLM agent to assist in reward design for RL \cite{ma2023eureka, xietext2reward}. While the synthetic reward signals can be used within a scripted teacher setup in PbRL, they may lack the historical context needed for nuanced evaluations due to the Markovian nature of the reward function \cite{kim2022preference}.}

\textcolor{black}{More importantly, the reasoning proficiency of an individual foundation agent}, characterized by inherent uncertainty and variability \cite{hong2023metagpt,chan2023chateval}, may not consistently match the expertise and reliability of human instructors. This challenge mirrors a familiar predicament in human society, where complex, multifaceted problems often surpass the expertise of any individual. In such situations, crowdsourcing has emerged as an efficient solution \cite{howe2006rise}. By harnessing the group wisdom of a distributed network of individuals, each contributing their unique knowledge and perspectives, crowdsourcing significantly enhances the utility of contributions from a vast pool of non-experts \cite{sheng2019machine}.

Drawing from this parallel, we introduce PrefCLM, a novel framework that leverages crowdsourced LLMs as simulated teachers in PbRL. As illustrated in Fig. \ref{fig:concept}, our approach centers on an LLM-based crowdsourced evaluation paradigm, which harnesses the diversity and collective intelligence of multiple LLM agents to evaluate robot behaviors. To efficiently manage the uncertainties and conflicts within the crowdsourcing, we further integrate the Dempster-Shafer Theory (DST) \cite{shafer1992dempster} to seamlessly fuse individual evaluative feedback from LLM instructors as a crowdsourced preference used in the PbRL process. Moreover, we introduce a human-in-the-loop (HITL) module that enables PrefCLM to incorporate language-based user feedback on iteratively learned robot policies during the PbRL process and accordingly adapt the evaluation patterns, leading to synthetic crowdsourced feedback that more closely aligns with unique user preferences in HRI scenarios. 

The overview of the proposed PrefCLM with four main components is depicted in Fig. \ref{fig:framwork}. Our key contributions in this letter can be summarized as:
\begin{itemize}[leftmargin=*]
    \item \textcolor{black}{We propose PrefCLM as a novel solution to enhancing the feedback efficiency in PbRL with synthetic preferences from crowdsourced LLM agents.} PrefCLM can also refine synthetic preferences to match unique and subtle user expectations more closely through interactive user inputs.
    \item We introduce crowdsourcing and DST principles to efficiently leverage the collective intelligence and variegation of multiple LLM agents in evaluating robot behaviors and reflecting user individual expectations, elevating the quality and relevance of the crowdsourced preference feedback.
    \item Our extensive experiments demonstrate that PrefCLM can achieve competitive or superior task performance compared to expert-tuned scripted teachers across various benchmark general RL tasks. As supported by a user study (N=10), PrefCLM with user interactive inputs can facilitate more personalized robot behaviors, significantly boosting user satisfaction in real-world HRI scenarios. 
\end{itemize}


\section{Background and Preliminary}
\label{PF}
\subsection{Preference-based RL and Scripted Teachers.}
PbRL has emerged as a promising approach to circumvent the complexities with reward function engineering in traditional RL \cite{lee2021pebble}. Its goal is to learn a preference-aligned reward model $\widehat{R}_\psi$, typically a neural network parameterized by $\psi$, based on comparative feedback between segments of robot trajectories (or behaviors) provided by a human instructor. 

A robot trajectory segment $\sigma$ is defined as a time-indexed sequence of states and actions within a specified length $L$: $\left\{\left(\mathbf{s}_1, \mathbf{a}_1\right), \ldots,\left(\mathbf{s}_L, \mathbf{a}_L\right)\right\}$. Then a human teacher is periodically asked to express their preference $\Lambda \in \{{0,1,0.5}\}$ between a pair of trajectory segments ${\sigma^0, \sigma^1}$, where $\Lambda=1$ indicates $\sigma^1$ is preferred, $\Lambda=0$ denotes the opposite, and $\Lambda=0.5$ means equally preferred. Each preference query is stored in a replay buffer $\mathcal{B}$ as $\left(\sigma^0, \sigma^1, \Lambda\right)$. Then, a Bradley-Terry model-based preference predictor \cite{christiano2017deep} is employed to estimate the preference probabilities as:
\begin{equation}
\mathcal{P}_\psi\left[\sigma^1 \succ \sigma^0\right]=\frac{\exp \left(\sum_t \widehat{R}_\psi\left(\mathbf{s}_t^1, \mathbf{a}_t^1\right)\right)}{\sum_{i \in\{0,1\}} \exp \left(\sum_t \widehat{R}_\psi\left(\mathbf{s}_t^i, \mathbf{a}_t^i\right)\right)}
\label{BT}
\end{equation}
\noindent where $\mathcal{P}_\psi\left[\sigma^1 \succ \sigma^0\right]$ represents the probability that trajectory segment $\sigma^1$ is preferred over $\sigma^0$.

The reward learning problem is then regarded as a supervised learning problem, with the objective of minimizing the binary cross-entropy loss between the preferences provided by humans $\Lambda$ and those by the reward model $\widehat{R}_\psi$ as:

\begin{equation}
\begin{split}
\mathcal{L}_{\psi}=-\sum_{\left(\sigma^{1}, \sigma^{0}, \Lambda \right) \in \mathcal{B}} &\Lambda(1) \log \mathcal{P}_{\psi}\left[\sigma^{1} \succ \sigma^{0}\right]+ \\ &\Lambda(0) \log \mathcal{P}_{\psi}\left[\sigma^{0} \succ \sigma^{1}\right]
\end{split}
\label{loss}
\end{equation}

\textcolor{black}{To enable easier and more systematic PbRL algorithm benchmarking}, most exisiting works \cite{christiano2017deep,lee2021pebble,lee2021b,park2021surf,zhao2024prefmmt,metcalf2023sample,liu2022task} have adopted a scripted teacher that provides synthetic preferences derived from an engineered reward function $R$ for training. Formally, a scripted teacher determines synthetic preferences by comparing the cumulative reward values of each trajectory as:
\begin{equation}
\Lambda= \begin{cases} 0 & \text { If } \sum_{t=1}^L R\left(\mathbf{s}_t^0, \mathbf{a}_t^0\right)>\sum_{t=1}^L R\left(\mathbf{s}_t^1, \mathbf{a}_t^1\right) \\ 1 & \text { otherwise }\end{cases}
\label{simulated_teacher}
\end{equation}

Incorporating the stochastic labeling strategies \cite{lee2021b}, a scripted teacher can offer more realistic feedback with irrationalities. \textcolor{black}{While beneficial for controlled testing, scripted teachers are not intended for real-world applications and are thus unsuitable for such settings:} their reliance on predefined reward functions limits flexibility and often prevents them from capturing long-term information within trajectories due to the static and Markovian nature of these functions \cite{kim2022preference}. In contrast, PrefCLM harnesses the collective intelligence of multiple LLMs to evaluate robot behaviors, providing a more nuanced and adaptable evaluation pattern.

\subsection{Foundation Agents for Synthetic Feedback in RL}
Recent works have explored the potentials of foundation agents for generating synthetic feedback in robot learning. For example, \cite{ma2023eureka, xietext2reward} employ a coding-LLM agent to design reward functions and generate synthetic reward signals in RL. Unlike these approaches, we prompt each LLM agent to generate non-Markovian evaluation functions that consider not only the current state-action pair but also contextual information across the entire trajectory in PbRL. \textcolor{black}{Another more related work \cite{wangrl} utilizes a VLM agent to provide synthetic preference labels in PbRL by analyzing high-level state information from image sequences of trajectory pairs. In contrast, our approach integrates both state and action information across trajectories with greater granularity, enabling a more comprehensive assessment that captures critical nuances such as action smoothness and intent.}

\textcolor{black}{More importantly, unlike these previous works that rely on a single foundation agent, we employ a crowdsourcing strategy and strategic fusion methods to harness the collective intelligence of multiple LLM agents. This approach enhances the quality of generated feedback and facilitates greater adaptability to reflect personalized user expectations in HRI.}


\begin{figure*}[!t]
\centering
\includegraphics[width=\linewidth]{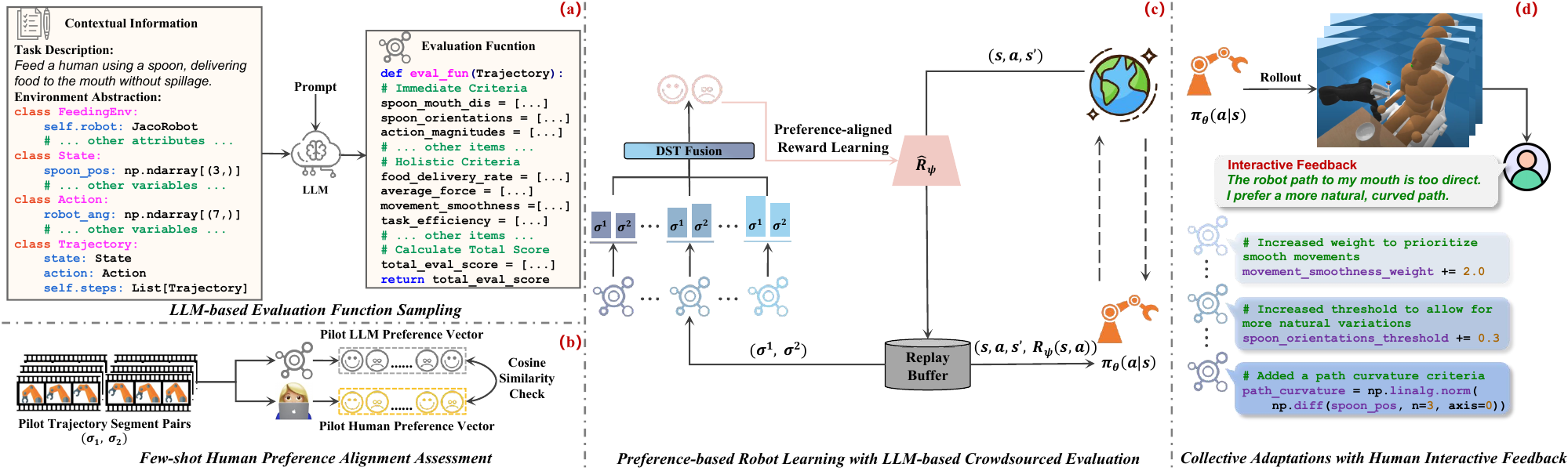}
\vspace{-20pt}
\caption{Overview of the PrefCLM framework. (a) Given task-specific contextual information and prompts, multiple code-based evaluation functions are sampled from crowd LLM agents (Section \ref{sec:function_sampling}). (b) A cosine similarity check module then filters the sampled evaluation functions, selecting those that align with few-shot expert preferences within a specified tolerance (optional, Section \ref{sec:function_filtering}). (c) Evaluative scores are continuously assigned by these selected evaluation functions to pairs of robot trajectories. These scores are aggregated through Dempster-Shafer Theory (DST) fusion to form crowdsourced preferences, which are used for the reward learning in PbRL (Section \ref{sec:preference_fusion}). (d)  Crowd LLM agents can also collectively adapt and refine their evaluation functions based on user interactive inputs given periodically in HRI scenarios (optional, Section \ref{sec:hri}).}
\vspace{-15pt}
\label{fig:framwork}
\end{figure*}
\subsection{Compound LLM Systems and Crowdsourcing.}
A single LLM agent often faces reasoning deficiencies, such as hallucinations and limited perspectives, making it less effective for complex tasks \cite{chan2023chateval}. To address this, compound LLM systems have been proposed, wherein multiple LLMs act in different roles to handle various aspects of the problem \cite{hong2023metagpt} or engage in debates to produce a more comprehensive output \cite{chan2023chateval}. While we follow this general approach, our method differs in that we incorporate the repeated labeling concept from crowdsourcing \cite{sheng2019machine}, where each LLM independently handles the same role of generating evaluation functions for robot trajectories. Furthermore, differing from the prevalent majority voting method \cite{sheng2019machine} or \textcolor{black}{spectral meta-learner approach \cite{chhan2024crowd} used to aggregate decisions (e.g., preferences) in compound systems or crowdsourcing, we employ DST \cite{shafer1992dempster} for integrating the outputs at the score level from each LLM (e.g., evaluative scores) in evaluating robot behaviors. This approach enables a more nuanced and informative fusion of preferences from the LLM crowd, taking into account the confidence of each preference and addressing associated indecision or conflicts more effectively.}


\section{Methodology}
In this section, we present PrefCLM, a framework utilizing crowdsourced LLM agents to enable zero-shot or few-shot synthetic preference generation within PbRL. Furthermore, we introduce a HITL module in PrefCLM that can incorporate human interactive inputs to dynamically adapt and refine the evaluation mechanisms, leading to more personalized robot behaviors in HRI scenarios.

\subsection{LLM-based Evaluation Function Sampling}
\label{sec:function_sampling}
Assuming $n$ LLM agents comprise the crowd, which can be either homogeneous or heterogeneous, the initial step in PrefCLM involves sampling multiple code-based evaluation functions from these agents. Each function, represented as $\mathcal{F}: \dot{\sigma} \rightarrow \dot{\rho}$, calculates an evaluative score $\dot{\rho}$ for a robot trajectory segment $\dot{\sigma}$ within a given task scenario. We prompt LLM agents to produce functions that regard the whole robot trajectory as the evaluative object, instead of single state-action pairs as considered in scripted teachers. Such functions aim to evaluate the holistic patterns and changes across time-steps within the entire trajectory in addition to the immediate effectiveness of each state-action pair, ensuring a more nuanced evaluation akin to humans. 

To this end, contextual information along with prompts $p$ (Appx. \ref{appendix:A}) is provided to LLM agents. \textcolor{black}{To minimize prompt engineering effort, we use a unified template as shown in Fig. \ref{fig:framwork}.} The contextual information includes a concise task description $l$, which delineates the objectives of the task (Appx. \ref{appendix:B}), and an environment abstraction $e$, structured as a hierarchy of Pythonic classes following \cite{xietext2reward} (Appx. \ref{appendix:C}).  This Pythonic representation provides a high-level and essential overview of the task environment, with particular emphasis on states, actions, and trajectories. Utilizing this information, the crowd LLM agents individually engage in-context reasoning to generate $n$ evaluation functions as:
\begin{equation}
\mathcal{F}_1, \ldots, \mathcal{F}_n \sim \operatorname{LLMs}(l, e, p)
\label{sampling}
\end{equation}

Empirically, we observe that the evaluation functions, even those generated by homogeneous agents, exhibit diversity (Appx. \ref{appendix:D}). This variation manifests in several ways, such as differing task-related criteria, assorted definitions for the same criteria, and varying priorities assigned to these criteria (e.g., different weighting schemes). Our PrefCLM capitalizes on this diversity, leveraging the unique perspectives that each LLM agent brings to the task, \textcolor{black}{resulting in a richer and more comprehensive evaluation process under the same level of prompting.}


\subsection{Few-shot Expert Preference Alignment}
\label{sec:function_filtering}
Few-shot expert involvement has proven to enhance the performance of LLM-based systems \cite{wu2023tidybot,singh2023progprompt}. In line with this, our method integrates a few-shot generation pattern in addition to the zero-shot evaluation function generation previously described. This is facilitated by a few-shot expert preference alignment module. Specifically, an expert preference vector, $\Upsilon_{expert} = [\Lambda^1, \dots, \Lambda^j]$, is constructed by a human expert giving preferences $\Lambda$ for a set of $j$ pilot trajectory segment pairs. Similarly, for each evaluation function $\mathcal{F}$ sampled, we can derive an LLM preference vector, $\Upsilon_{\mathcal{F}} = [\dot{\Lambda}^1, \dots, \dot{\Lambda}^j]$, where $\dot{\Lambda}$ represents a pseudo preference for a segment pair, determined by the relative magnitude of evaluative scores that $\mathcal{F}$ assigns to each segment in the pair. Then we can calculate the cosine similarity score, denoted as $\vartheta \in [-1,1]$, between the $\Upsilon_{expert}$ and $\Upsilon_{\mathcal{F}}$ as: 
\begin{equation}
\vartheta =\frac{\Upsilon_{{expert }} \cdot \Upsilon_{\mathcal{F}}}{\left\|\Upsilon_{ {expert }}\right\| \times\left\|\Upsilon_{\mathcal{F}}\right\|}
\end{equation}

Cosine similarity quantifies the angular closeness between these two vectors irrespective of their magnitude, effectively capturing the alignment in preference orientation between the expert and the LLM agent in high-dimensional spaces. Evaluation functions that achieve a similarity score exceeding a threshold $\hat{\vartheta}$ are selected for further use. This ensures that selected functions are closely aligned with expert preferences while maintaining a diverse set of evaluative perspectives.

\subsection{Crowdsourced Evaluation and Preference Fusion}
\label{sec:preference_fusion}
The subsequent phase entails employing the zero-shot generated or few-shot selected evaluation functions for the crowdsourced evaluation. This involves aggregating individual preferences $\dot{\Lambda}$ from $n$ crowd LLM agents to form a unified crowdsourced preference $\overline{\Lambda}$, which is then utilized for the reward learning process as desceibed in Eqs. \ref{BT} and \ref{loss}. For this purpose, we adopt DST \cite{shafer1992dempster} as the fusion strategy at the level of evaluative scores. 

Consider a pair of robot trajectory segments, $\sigma^0$ and $\sigma^1$, to be evaluated by the LLM crowdsourcing. We define the frame of discernment in DST as $\Theta= \{\sigma^0, \sigma^1, \{\sigma^0,\sigma^1\}\}$, which represents the possible decisions of preferring $\sigma^0$, preferring $\sigma^1$, or having an equal or indeterminate preference between them, respectively. For the $k^{th}$ LLM agent, we first normalize its assigned scores $\rho^0_k$ and $\rho^1_k$ for $\sigma^0$ and $\sigma^1$ as: 
\begin{equation}
\hat{\rho}^0_k = \frac{\rho^0_k}{\rho^0_k + \rho^1_k}; \hat{\rho}^1_k = \frac{\rho^1_k}{\rho^0_k + \rho^1_k}
\label{norm}
\end{equation}

Normalization is crucial here since the scale of the evaluative score from each agent is unknown and variable. We then calculate the mass function, which reflects the $k^{th}$ agent's belief toward each decision in $\Theta$ as:

\begin{equation}
\begin{split}
m_k(\{\sigma^0, \sigma^1\}) &= \varphi \times \left(1 - |\hat{\rho}^0_k - \hat{\rho}^1_k|\right)\\
m_k(\sigma^0) = \hat{\rho}^0_k &\times (1 - m_k(\{\sigma^0, \sigma^1\})) \\
m_k(\sigma^1) = \hat{\rho}^1_k &\times (1 - m_k(\{\sigma^0, \sigma^1\}))
\end{split}
\end{equation}

Here, $\varphi \in [0,1]$ is a parameter in DST that determines the maximum level of indecision assignable by any agent.
Then the fused mass function for any decision $X$ within $\Theta$ is calculated as:
\begin{equation}
\overline{m}(X)=\bigoplus_{k=1}^n m_k(X)
\end{equation}
where $\bigoplus$ denotes the operation of Dempster’s rule of combination that fuses individual belief iteratively as:
\begin{equation}
m_{1: k}(X)=\frac{1}{1-K_{1: k}} \sum_{A \cap B=X} m_{1: k-1}(A) \cdot m_k(B)
\end{equation}
\begin{equation}
K_{1: k}=\sum_{A \cap B=\emptyset} m_{1: k-1}(A) \cdot m_k(B)
\end{equation}
where $K_{1: k}$ represents the conflict measure after combining up to the $k^{th}$ LLM agent. $A$ and $B$ are subsets of the possible preference decisions within $\Theta$.

Subsequently, the decision in $\Theta$ with the highest aggregated belief after considering all $n$ LLM agents is determined as the final decision, $\overline{X}$, as:
\begin{equation}
\overline{X} = \underset{X \in \Theta}{\operatorname{argmax}} \overline{m}(X)
\end{equation} 

Finally, we can obtain the crowdsourced preference $\overline{\Lambda}$ as:
\begin{equation}
\overline{\Lambda} =
\begin{cases}
0 & \text{if } \overline{X} = \sigma^0 \\
1 & \text{if } \overline{X} = \sigma^1 \\
0.5 & \text{otherwise} 
\end{cases}
\end{equation}

\textcolor{black}{Through the fusion procedures of individual preference beliefs described above, we effectively leverage the diversity of the LLM crowd while mitigating associated downsides, such as increased potential uncertainty and conflicts.} The crowdsourced preference is generated each time the PbRL algorithm queries for a preference between a pair of trajectory segments rolled out from the current robot policy. This synthetic preference adheres to the same format as those obtained from human teachers, denoted as $\Lambda$ in Section \ref{PF}. As a result, the PrefCLM framework can be seamlessly integrated into the reward learning phase in any PbRL frameworks, offering a plug-and-play solution, eliminating the need for extensive human efforts in providing feedback.

\subsection{Collective Adaptations with Interactive Inputs for HRI}
\label{sec:hri}
The crowdsourced preference aggregated from crowd LLM agents can be semantically consistent with the objectives of general RL tasks, \textcolor{black}{yet it may be \textit{misidentified} in practice \cite{tien2023causal},} especially in HRI scenarios where expectations of the same task may vary across users \cite{wang2024personalization}. 

To better adapt to such unique user expectations, we propose an extension of PrefCLM tailored for HRI scenarios. The first intuitive step is to incorporate specific user expectations, such as ``I prefer the robot to move slowly and cautiously when feeding me'', into the prompts for the evaluation function sampling stage in PrefCLM as additional contextual information. This allows LLM agents to integrate specific evaluation criteria that reflect user preferences, such as motion caution, into their evaluation functions. 


However, humans rarely articulate their intentions clearly or comprehensively in a single or few instances, and their expectations may change as the interaction evolves \cite{gasteiger2023factors}. To address this, as illustrated in Fig. \ref{fig:framwork}d, we introduce an additional HITL module in PrefCLM capable of dynamically incorporating interactive inputs from users into the evaluation paradigms of the LLM crowd, thus fostering more personalized robot behaviors. In addition to initializing PrefCLM with user expectations as described above, users can observe the robot trajectory segments rolled out from the iterative robot policies during training sessions and offer real-time interactive feedback, such as ``The robot's path to my mouth is too direct. I prefer a more natural, curved path.''

This interactive feedback is conveyed to the crowd of LLM agents, triggering a collective refinement of their evaluation functions to better align with the given user preferences. Each LLM agent may interpret and incorporate the user feedback differently based on their current evaluation functions. For instance, as shown in Fig. \ref{fig:framwork}d, one agent might focus on adjusting existing related criteria, such as increasing the importance of \textit{motion smoothness} to encourage gentle and continuous movements or modifying the threshold for acceptable \textit{end-effector orientation} to allow for more variations in the angle of the spoon. Another agent might introduce a new evaluation criterion, such as \textit{path curvature}, which assigns higher scores to paths that exhibit a more fluid and curved shape, if such criteria were previously lacking in their evaluation function. This diversity in adjustments with the DST fusion could ensure that the refined pattern of the LLM-based crowdsourced evaluation capture a broad spectrum of the given interactive feedback.

The updated evaluation functions are then used to re-score the robot trajectory segments $\sigma$, both those stored in the replay buffer and newly generated ones, producing new preference labels $\overline{\Lambda}$ that better align with the expressed feedback of the user. These updated preference labels are subsequently used to retrain the reward model $\widehat{R}_\psi$, which in turn guides the robot learning process towards generating behaviors that more closely match the user expectations.

\section{Experiments and Results}
\label{exp}
\subsection{Experiments on General RL Tasks}
We first designed experiments to evaluate the performance of PrefCLM under zero-shot and few-shot generation conditions on general RL tasks across various simulation benchmarks. Extensive ablation studies were also conducted.
\subsubsection{Environments} 
The general RL tasks we consider include locomotion tasks from the DeepMind Control Suite benchmark \cite{tassa2018deepmind}, specifically the Walker, Cheetah, and Quadruped tasks, as well as manipulation tasks from the Meta-World benchmark \cite{yu2020meta}, specifically the Button Press, Door Unlock, and Drawer Open tasks. 

\subsubsection{Baselines}
We compared our method against two baselines for generating synthetic preferences in PbRL: 

\begin{itemize}[leftmargin=*]
    \item Scripted Teachers. This baseline determines preferences towards robot trajectories based on expert-tuned reward functions, representing a common practice in existing PbRL works \cite{christiano2017deep,lee2021pebble,lee2021b,park2021surf,zhao2024prefmmt,metcalf2023sample,liu2022task}. 
    \item PrefEVO. We built this baseline by adapting the Eureka framework \cite{ma2023eureka}, initially designed for reward design with an LLM agent, to PbRL scenarios. The framework uses evolutionary search to enhance LLM outputs by sampling and evaluating multiple reward functions, selecting the one achieving highest scores from pre-defined fitness functions as the final reward function. The LLM can also refine the final reward function by self-reflecting on the policy performance metrics, e.g., the success rate and its changes on manipulation tasks. 
    
    ~~To adapt it to PbRL settings, we replace the prompts and contextual information with those used in PrefCLM, and substitute the fitness functions with the cosine similarity scores mentioned in Section \ref{sec:function_filtering} as the search criterion, since such functions are not available in our setup. This represents a state-of-the-art single-LLM-based baseline.
\end{itemize}
\textcolor{black}{We also tested PrefCLM and PrefEVO with the HITL module on general RL tasks. The results and analysis are provided in Appx. \ref{appendix:F}.}

\subsubsection{Ablation Studies}
To further investigate the impact of crowdsourcing and DST fusion mechanisms within our framework, we conducted additional ablation studies. Specifically, we aimed to assess how the number and composition (homogeneous or heterogeneous) of the LLM agents in the crowdsourcing affect the performance, and how DST fusion benefits PrefCLM under these conditions. 

To this end, we implemented an ablation model named MajCLM, which uses majority voting, a common method in compound AI systems, to fuse individual preferences from crowd LLM agents instead of using DST fusion. We tested PrefCLM and MajCLM on the Walker and Button Press tasks with different numbers of homogeneous LLM agents ($3,10,20$) under zero-shot settings, using GPT-4 \cite{achiam2023gpt} as the LLM backbone. Results of a single LLM agent are also reported. We further tested them under heterogeneous settings with a combination of GPT-4, Claude-Opus, and Llama-3-70B.
\textcolor{black}{We also conducted two ablation studies on the backbone PbRL algorithm and the composition and quality of LLMs, as detailed in Appx. \ref{appendix:F}.}
\subsubsection{Implementation Details}
All methods were implemented using PEBBLE \cite{lee2021pebble}, a benchmark PbRL algorithm, as the backbone network with consistent hyperparameters, and disagreement sampling was used for feedback querying. Feedback volume was determined in line with prior research \cite{liu2022meta, park2021surf}, assigning $600$ queries for tasks such as Walker and Cheetah, $1200$ for Quadruped, $2500$ for Button Press and Door Unlock, and $5000$ for Drawer Open. 

For the Scripted Teachers baseline, benchmark-provided reward functions were used, and we adopted the optimal labeling strategies for each task as reported in \cite{lee2021b}. For all LLM-based methods, GPT-4, specifically the \texttt{gpt-4o} model, was employed as the LLM backbone, unless stated otherwise. The crowd size for PrefCLM was set at $10$ when compared to other baselines. In implementing the few-shot expert preference alignment in PrefCLM and the evolutionary search in PrefEVO, we provided $15$ pilot preferences for trajectories rolled out from the policy during the unsupervised learning phase in PEBBLE for each task. For PrefEVO, following the original work \cite{ma2023eureka}, we conducted $5$ rounds of evolutionary search with a sampling size of $16$ each time. The threshold $\hat{\vartheta}$ of cosine similarly in PrefCLM was set to $0.5$ for locomotion tasks and $0.6$ for manipulation tasks. Additionally, the indecision parameter $\varphi$ in DST was set to $0.3$ for all tasks. 

Following \cite{liu2022meta}, we used ground-truth reward returns as the performance metric for locomotion tasks, while for manipulation tasks, we reported the success rate as provided by the benchmark. Each model underwent five independent runs on each task, with results reported as an average accompanied by the standard deviation. All experiments were conducted on a workstation with three NVIDIA RTX 4090 GPUs.

\begin{figure*}[!t]
\centering
\includegraphics[width=\linewidth]{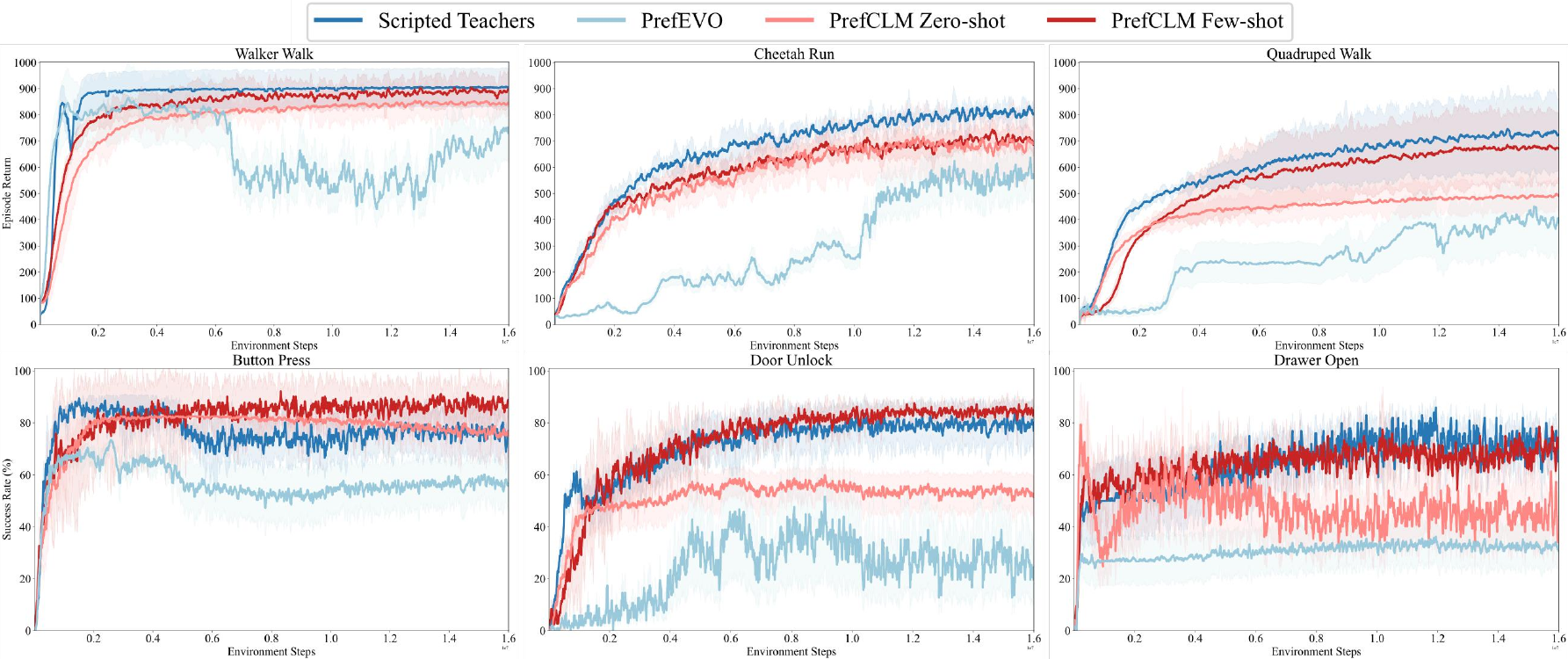}
\vspace{-20pt}
\caption{Learning curves on general RL tasks, measured in episode returns for locomotion tasks and success rates for manipulation tasks. The solid line represents the mean, while the shaded area indicates the standard deviation across five runs.}
\vspace{-15pt}
\label{fig:result1}
\end{figure*}

\begin{figure}[!t]
\centering
\includegraphics[width=\linewidth]{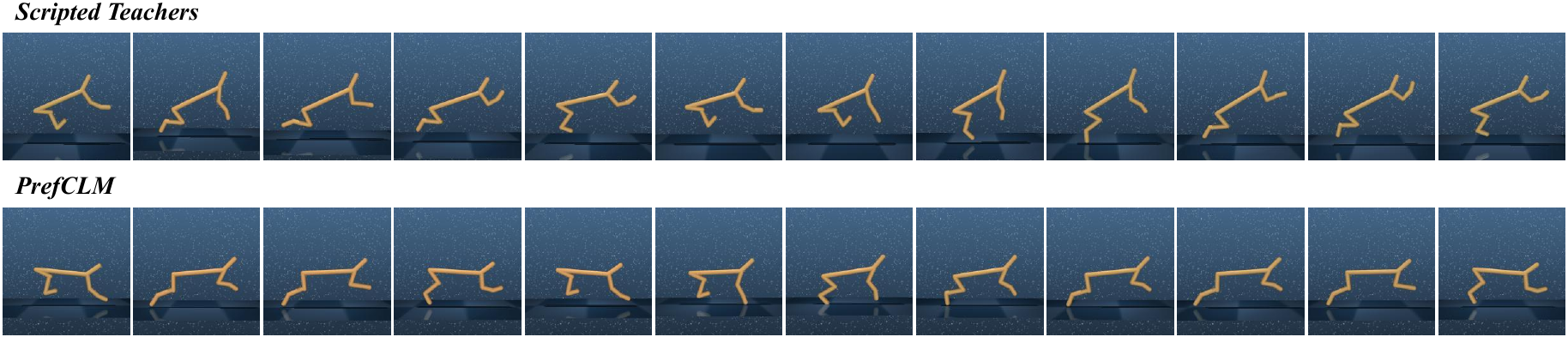}
\vspace{-20pt}
\caption{Locomotion behaviors learned by the Scripted Teachers (top) and PrefCLM (bottom) on the Cheetah Run task.}
\vspace{-20pt}
\label{fig:visual}
\end{figure}

\subsubsection{Results and Analysis}
Figure \ref{fig:result1} illustrates the learning curves of PrefCLM in comparison to various baselines. We can observe that PrefCLM, under zero-shot or few-shot generation modes, achieves performance comparable to expert-tuned Scripted Teachers across most locomotion and manipulation tasks in terms of episode returns or final success rates and convergence speed. Notably, PrefCLM even outperforms the baseline on the Button Press and Door Unlock tasks. Although PrefCLM does not surpass the Scripted Teachers in locomotion tasks, it leads to more natural and efficient behaviors, as illustrated in Figure \ref{fig:visual}, where the robot behaviors from PrefCLM use two legs to run like a real animal, whereas those from Scripted Teachers only use the back leg to jump. This advantage is attributed to the LLM-generated evaluation functions which consider a broader range of criteria, such as motion efficiency, in addition to the basic success criteria typically considered by reward functions in Scripted Teachers. Additionally, the Markovian nature of reward functions in Scripted Teachers may neglect historical information that is critical for capturing the full context of robot behaviors. Overall, these observations suggest that PrefCLM can generate high-quality and more comprehensive evaluation patterns without the need for expert engineering, showcasing its potential as an efficient simulated teacher in PbRL.

\begin{figure}[!t]
\centering
\includegraphics[width=\linewidth]{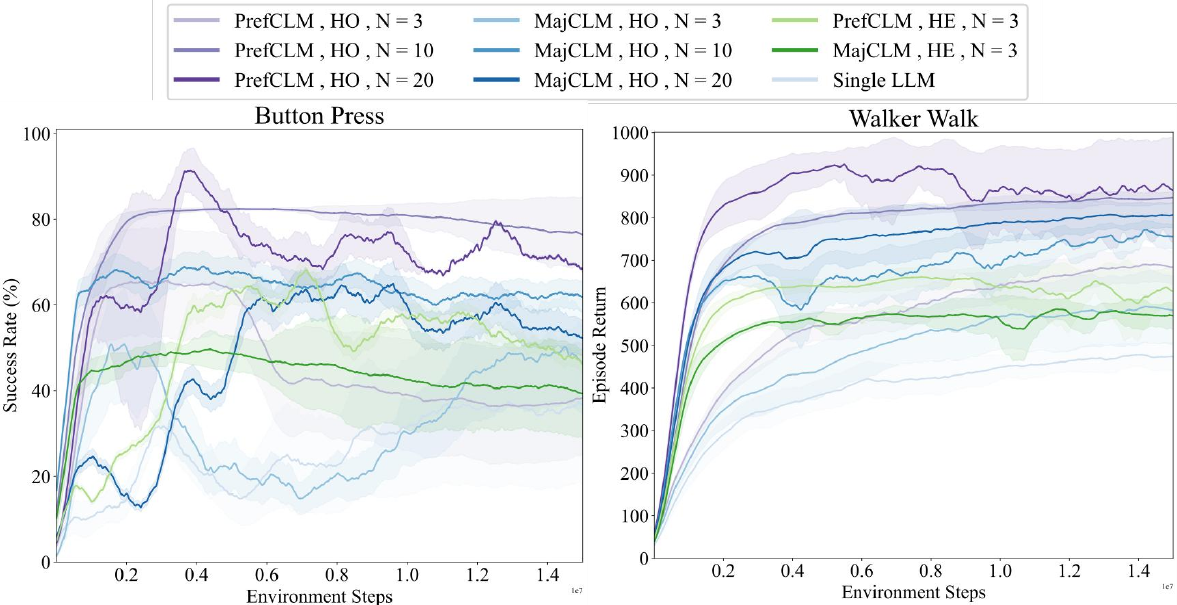}
\vspace{-20pt}
\caption{Results of ablation studies in terms of learning curves \textcolor{black}{with a moving window average of 100 applied for readability}. HO: homogeneous setting; HE: heterogeneous setting; N: number of LLM agents in the crowd.}
\vspace{-20pt}
\label{fig:ablation}
\end{figure}

Furthermore, while the zero-shot mode of PrefCLM can achieve satisfactory results, a notable performance gap is observed in incompletely solved tasks such as Quadruped and Drawer Open. This highlights the benefits of the few-shot expert alignment process, especially when the task demands nuanced understanding.

On the other hand, we note a significant performance disparity between PrefEVO and PrefCLM, with PrefEVO failing to surpass even the zero-shot mode of PrefCLM. This discrepancy can be attributed to the limitations of the evolutionary search method in the context of PbRL. Unlike traditional RL, where reward outputs directly dictate robot policy, PbRL involves a two-tiered process: evaluative scores are first used to learn a reward model, which then informs the policy. Consequently, the self-reflection metrics employed in PrefEVO may not effectively represent the quality of the evaluative functions, as these metrics are not directly linked to the overall effectiveness of the feedback mechanism in PbRL. Therefore, the evolved evaluation functions may not translate into improved performance in the PbRL setting. Additionally, the time and resource-intensive nature of evolutionary search makes it less efficient. These demonstrate that the crowdsourcing approach in PrefCLM is a more fitting and efficient method for enhancing LLM-based evaluations in PbRL, leveraging the collective wisdom of crowd LLMs.

Moreover, as depicted in Fig. \ref{fig:ablation}, we observe that the PrefCLM (purples) outperforms MajCLM (blues) under different crowd numbers in homogeneous settings and in the heterogeneous setting. This can be attributed to the fact that the majority voting fusion method in MajCLM may not efficiently handle conflicts and uncertainties, which increase as the size and heterogeneity of the crowd agents grow. Majority voting at the decision level may overlook the nuances and disagreements among the agents. In contrast, the DST fusion in PrefCLM effectively manages these complexities by handling the diverse beliefs and resolving conflicts at the score level, offering a more robust fusion mechanism.

Additionally, as shown in Fig. \ref{fig:ablation}, we can observe the effects of crowd LLM agent size. For both models, performance generally increases with the crowd size. However, this performance enhancement diminishes as the number continues to grow. For example, moving from n=1 (single LLM) to n=3 and then to n=10 shows continuous improvement, but the differences between n=10 and n=20 are minimal, with performance remaining nearly the same. This trend aligns with findings in compound inference systems \cite{chen2024more}.

\begin{figure}[t]
\centering
\includegraphics[width=0.9\linewidth]{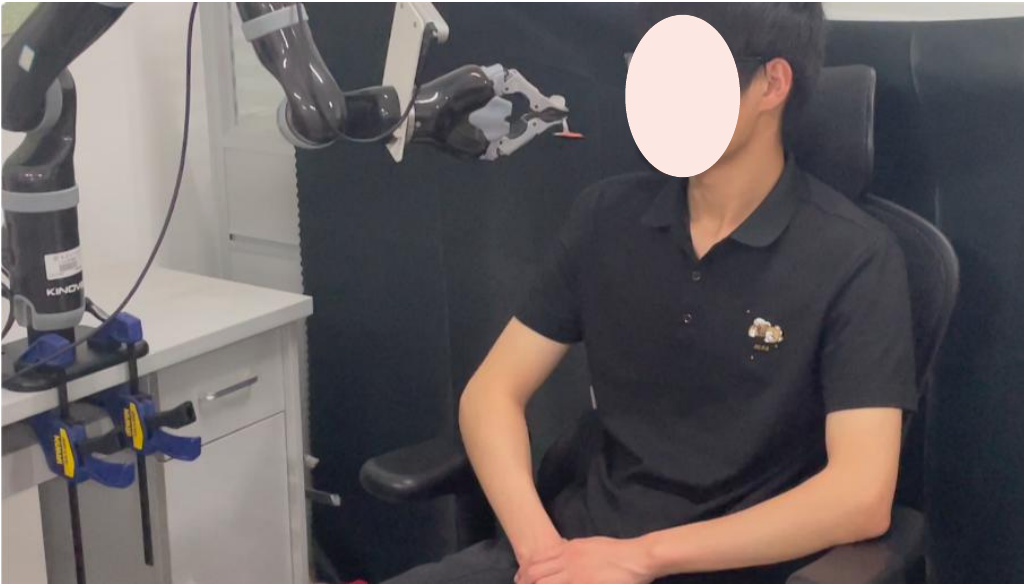}
\vspace{-10pt}
\caption{Illustration of the user study environment. A participant seated in a chair interacts with a Jaco assistive robot arm that feeds using a spoon.}
\vspace{-20pt}
\label{fig:env}
\end{figure}

\subsection{Real-world User Study}
To evaluate the personalization and user satisfaction enabled by PrefCLM with the HITL module in realistic HRI tasks, we further carried out a real-world user study.

\subsubsection{Setup} The HRI task we selected is the Feeding task from the Assistive Gym benchmark \cite{erickson2020assistive}. In this task, a robot arm is tasked with delivering a spoon holding food, represented as small spheres, to the mouth of a human seated in a chair without spilling. For the real-world setup, as illustrated in Fig. \ref{fig:env}, we employed the Kinova Jaco assistive robotic arm and a RealSense D435 camera, integrated with a face landmark detection algorithm \cite{lugaresi2019mediapipe} and YOLOv7 to track the human's head and mouth and the position of the spoon respectively. An emergency switch is conveniently located and easily accessible to investigators, ensuring safety throughout the experiment. 

\subsubsection{Baselines} To guarantee the safety of participants in our user study, we initially pre-trained a policy following the guidance from the benchmark \cite{erickson2020assistive} in simulation. This step is to ensure the basic functionality, i.e., successfully reaching the human mouth. Additionally, we configured the simulation environment to closely mimic the real-world setup, including the position of the chair, human shape, and the mounting of the robotic arm, in order to minimize the sim-to-real gap. Following this, we fine-tuned the policy using synthetic feedback from PrefCLM (few-shot, n=10) equipped with the HITL module. Additionally, PrefEVO, also equipped with the same module, served as another baseline representing state-of-the-art single-LLM-based HITL adaptation approaches \cite{xietext2reward}.

\begin{figure}[t]
\centering
\includegraphics[width=\linewidth]{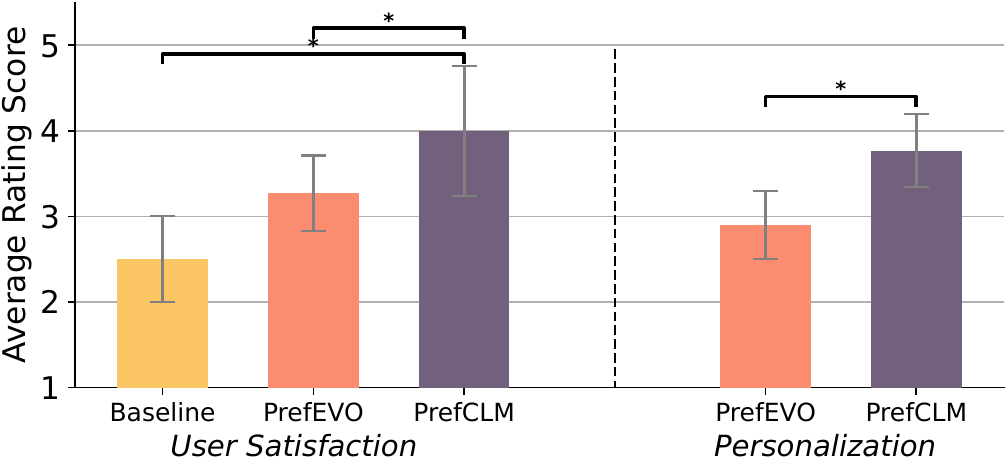}
\vspace{-20pt}
\caption{Average rating scores from participants in terms of satisfaction and personalization. Basline refers to the pre-trained policy. *:$p<0.01$.}
\vspace{-20pt}
\label{fig:user}
\end{figure}

\subsubsection{Experimental Design}
We recruited 10 participants (3 females, 7 males) from faculty/staff and students at BUCT, with an average age of 24.5 years (SD = 2.78). All participants reviewed and signed a consent form, which outlined the study objectives, procedures, potential risks, and their rights as participants. Initially, participants were introduced to the Feeding task and asked to verbally express their general expectations. These expectations, along with other contextual information and prompts, were used to generate the initial evaluation functions by PrefCLM and PrefEVO for the initial fine-tuning training of the pre-trained policy respectively. For each model, we periodically (every $4 \times 10^6$ environment steps, approximately 2 hours) rolled out the learned robot policy to the physical Kinova Jaco robotic arm, for a total of three times. Each time, each participant provided their interactive feedback, which was utilized to refine the evaluation functions.

After all the training, participants engaged in physical interactions with the robot policies fine-tuned by PrefCLM and PrefEVO, as well as with the pre-trained baseline policy. Each participant interacted with each policy three times in a randomized sequence and was not informed about which policy was active to prevent any bias in their responses. Following each interaction, participants were asked to rate the robot behaviors of the three policies in terms of satisfaction, using a Likert scale ranging from 1 (strongly disagree) to 5 (strongly agree). They were also asked to rate the level of personalization resulting from PrefCLM and PrefEVO using the same scale. To evaluate statistical differences between our model and others, two-sample t-tests were conducted.

\subsubsection{Results and Anslysis}
As shown in Fig \ref{fig:user}, we can observe that the policies fine-tuned by PrefEVO and PrefCLM achieve higher ratings in terms of overall user satisfaction compared to the pre-trained policy. This demonstrates that static reward functions, which serve as the basis of Scripted Teachers, lack the flexibility to capture subtle and unique user expectations in customizable HRI tasks. In contrast, the LLM-based approaches can more efficiently adapt to user expectations by incorporating interactive user verbal inputs and adjusting evaluation patterns in PbRL accordingly. This highlights the potential of LLM agents to bridge the gap between human intentions and robot behaviors.

More importantly, PrefCLM significantly outperforms PrefEVO in both satisfaction and personalization. This can be attributed to the fact that PrefEVO relies on the adaptation of the evaluation function by a single LLM agent, whereas PrefCLM adopts a collective adaptation pattern within the crowdsourcing framework. Each LLM agent refines its evaluation criteria based on user interactive inputs, covering a broader spectrum of user-specific feedback. \textcolor{black}{As one participant noted, ``\textit{I prefer that one (PrefCLM) because it made the robot's behavior feel more aligned with what I wanted after each of my comments.}" This highlights how the crowdsourcing and fusion strategies in our PrefCLM effectively harness collective intelligence to better align robot behaviors with individual user preferences.}

\section{Conclusion}
In this letter, we present PrefCLM that leverages crowdsourced LLMs for generating synthetic feedback in PbRL. PrefCLM is also capable of capturing the subtle intentions of users through interactive feedback. Our experimental results demonstrate that PrefCLM not only competes with expert-tuned scripted teachers in terms of performance in general RL tasks but also offers a more natural and intuitive method for specifying preferences in personalized HRI scenarios. \textcolor{black}{Future work could explore avenues such as controlling LLM agent diversity through intra-agent communication, investigating different fusion strategies, and extending to direct real-world observations in robot deployments.}

\typeout{}
\bibliography{main}
\bibliographystyle{IEEEtran}

\newpage

\onecolumn
\appendices

\section*{\textbf{\LARGE APPENDIX}}

\noindent In the appendix, we provide details about various aspects of our work, including the prompts used for generating evaluation functions, details of task descriptions and environment abstractions, example evaluation functions for the general RL tasks, and details about the pre-training and fine-tuning process in our user study.

\section*{\textbf{Table of Contents}}

I Prompts Used for Generating Evaluation Functions  \dotfill 1
\begin{itemize}
    \item I.1 Initial System Prompt
    \item I.2 Initial User Prompt
    \item I.3 Code Error Feedback Prompt
\end{itemize}

II Task Descriptions \dotfill 2
\begin{itemize}
    \item II.1 Description of Tasks Used in Benchmark
\end{itemize}

III Environment Abstraction \dotfill 2
\begin{itemize}
    \item III.1 Example Environment Abstraction for Walker Task
    \item III.2 Example Environment Abstraction for Button Press Task
    \item \textcolor{black}{III.3 Example Environment Abstraction for the Feeding Task}
\end{itemize}

IV Example of LLM-based Evaluation Functions \dotfill 5
\begin{itemize}
    \item IV.1 Expert-engineered Reward Function for Walker Task
    \item IV.2 LLM-based Evaluation Functions for Walker Task
    \item IV.3 Expert-engineered Reward Function for Button Press Task
    \item IV.4 LLM-based Evaluation Functions for Button Press Task
    \item IV.5 \textcolor{black}{LLM-based Evaluation Functions for the Feeding Jaco Task}
\end{itemize}

V The Feeding Task in the User Study \dotfill 17
\begin{itemize}
    \item V.1 Pre-trained Robot Policy
    \item V.2 Fine-tuning Process
\end{itemize}

\textcolor{black}{VI Additional Experimental Results} \dotfill 18
\begin{itemize}
\item \textcolor{black}{VI.1 Numerical Results of the Ablation Study on Agent Configurations (Fig. 5).}

    \item \textcolor{black}{VI.2 Experiments on General RL Tasks with HITL Module}
    \item \textcolor{black}{VI.3 Ablation Study on the Backbone PbRL Algorithm}
    \item \textcolor{black}{VI.4 Ablation Study on the Crowd Composition}
\end{itemize}

\section{Prompts Used for Generating Evaluation Functions} \label{appendix:A}
In this section, we present the prompts utilized for LLM-based evaluation functions sampling in the PrefCLM. We begin by initializing the LLM agent with a specific role and job description:

\begin{tcolorbox}[colback=lightlightgray, colframe=lightblue, title=\textbf{Prompt 1: Initial System}]
\texttt{ You are an expert evaluator specializing in preference-based reinforcement learning for robots. Your task is to design a sophisticated Python evaluation function that accurately scores robot trajectories within a specific reinforcement learning environment. This function is critical for guiding the robot's learning process and optimizing its task performance.}

\texttt{Your evaluation function should:}
\begin{itemize}
    \item \texttt{Use only the variables available in the robot's trajectory, which consists of multiple state-action pairs across different time steps.}
    \item \texttt{Return a single float value as the overall score, where higher scores indicate better performance.}
    \item \texttt{Incorporate two key components:}
    \begin{itemize}
        \item \texttt{\textit{Immediate evaluation:} Assess individual state-action pairs at each time step.}
        \item \texttt{\textit{Holistic evaluation:} Analyze patterns and trends across the entire trajectory.}
    \end{itemize}
\end{itemize}
\end{tcolorbox}

Next, we provide specific contextual information, including the task description (Appendix \ref{appendix:B}) and environment abstraction (Appendix \ref{appendix:C}), along with additional requirements for generating evaluation functions:

\begin{tcolorbox}[colback=lightlightgray, colframe=lightblue, title=\textbf{Prompt 2: Initial User}, breakable]
\texttt{I need you to generate the evaluation function for the following task: \{\textit{Task Description}\}. The Pythonic class-like environment abstraction is \{\textit{Environment Abstraction}\}.}

\texttt{Proceed as follows:}
\begin{itemize}
    \item \texttt{Analyze the task requirements and environment step-by-step.}
    \item \texttt{Develop a function with the signature \texttt{def evaluate\_trajectory(trajectory: Trajectory) -> float} that returns only the \texttt{final\_score}.}
    \item \texttt{Include comments in your code to explain your reasoning and design choices.}
\end{itemize}

\texttt{Additional Requirements:}
\begin{itemize}
    \item \texttt{The evaluation function must be a standalone function, suitable for integration into a class in another Python file.}
    \item \texttt{It must not contain any intra-class calls.}
    \item \texttt{Provide concrete, well-reasoned initial threshold values and weights. Avoid placeholders.}
\end{itemize}

\end{tcolorbox}

Additionally, in practice, although not frequent, sometimes the LLM agent may generate code with errors such as syntax errors or runtime issues (e.g., shape mismatch). In line with previous works \cite{ma2023eureka,xietext2reward}, we utilize the traceback message from code execution to prompt the LLM agent to fix the bug and provide an executable evaluation function if errors occur. The prompt for handling code errors is shown below:

\begin{tcolorbox}[colback=lightlightgray, colframe=lightblue, title=\textbf{Prompt 3: Code Error Feedback}]
\texttt{Executing the evaluation function code you generated bove has the following error: \{traceback_msg\}. Please fix the bug and provide a new, evaluation function.}

\end{tcolorbox}

\section{Task Descriptions} \label{appendix:B}
Following \cite{ma2023eureka,xietext2reward}, we use the task descriptions provided by the benchmark environments as the \{\textit{Task Description}\} in the prompts. These are summarized in Table.\ref{table:Task_description}.
\begin{table}[h]
\centering
\caption{Description of each task.}
\label{table:Task_description}
\begin{tabular}{c|c}
\hline
Task & Descriptions    \\ \hline\hline
                    
Walker Walk     &   Control the Walker robot to walk steadily in the forward direction, maintaining balance and speed.              \\ 
Cheetah Run     &   Control the Cheetah robot to run swiftly in the forward direction, optimizing for speed and stability.              \\ 
Quadruped Walk  &   Control the Quadruped robot to walk in the forward direction, ensuring coordination among all four legs for smooth movement.              \\ \hline\hline

Button Press    &   Instruct the robot to press a button located along the y-axis, requiring precise positioning and force application.\\
Door Unlock     &   Instruct the robot to unlock a door by rotating the lock mechanism counter-clockwise, requiring fine motor skills and dexterity.\\
Drawer Open     &   Instruct the robot to open a drawer by pulling its handle, requiring a firm grip and controlled pulling force. \\ \hline

\end{tabular}
\end{table}

\section{Environment Abstraction} \label{appendix:C}
To effectively generate evaluation functions within a task environment, LLM agents must understand how attributes of the robot and environment are represented, including the configuration of robots and objects, trajectory information, and available functions. To this end, following \cite{xietext2reward}, we employ a compact representation in Pythonic style, which utilizes Python classes, typing, and comments. This approach offers a higher level of abstraction compared to listing all environment-specific information in a list or table format, enabling the creation of general, reusable prompts across different environments. Additionally, Pythonic representation is prevalent in the pre-training data of LLMs, facilitating the LLM's understanding of the environment. Example environment abstractions for the Walker and Button Press tasks are provided below.

\begin{tcolorbox}[colback=lightlightgray, colframe=lightblue, title=Example Environment Abstraction for the Walker Task, breakable]
\begin{lstlisting}
class WalkerEnv:
    physics: Physics
    task: PlanarWalker
    control_timestep: float = 0.025  # Time interval for each control update.
    time_limit: float = 25  # Maximum duration for each episode in seconds.

    def step(self, action: np.ndarray):
        """Executes one timestep of the environment's dynamics with the given action and updates the trajectory."""
        pass

    def reset(self):
        """Resets the environment to an initial state and returns the first observation."""
        pass

    def get_trajectory(self) -> Trajectory:
        """Returns the trajectory data collected during an episode, including states, actions, and observations."""
        pass

class Physics:
    def torso_upright(self) -> float:
        """Calculates the cosine of the angle between the torso's z-axis and the vertical,
        indicating how upright the torso is."""
        pass

    def torso_height(self) -> float:
        """Returns the vertical position of the torso in meters, which helps monitor the walker's balance."""
        pass

    def horizontal_velocity(self) -> float:
        """Measures the horizontal speed of the walker's center of mass, reflecting movement efficiency."""
        pass

    def orientations(self) -> np.ndarray:
        """Returns an array of planar orientations for body segments, aiding in posture analysis."""
        pass

    def velocity(self) -> np.ndarray:
        """Returns a comprehensive velocity vector for all body parts, including both linear and angular velocities."""
        pass

class PlanarWalker:
    trajectory: Trajectory
    _move_speed: float  # Desired movement speed, varies with the task ('stand', 'walk', 'run').

    def get_observation(self, physics: Physics) -> collections.OrderedDict:
        """Compiles observational data from physics simulations, crucial for real-time decision-making."""
        pass

    def get_state(self, physics: Physics) -> dict:
        """Aggregates current state information from physics, providing a detailed snapshot of dynamic conditions."""
        pass

class Trajectory:
    def __init__(self, max_length=time_limit):
        self.states: deque  # queue of states, max length 25
        self.actions: deque  # queue of actions, max length 25
        self.observations: deque  # queue of observations, max length 25

    def add_step(self, state: dict, action: np.ndarray, observation: np.ndarray):
        # Add a step to the trajectory

    def __len__(self) -> int:
        # Return the number of steps in the trajectory
\end{lstlisting}
\end{tcolorbox}

\begin{tcolorbox}[colback=lightlightgray, colframe=lightblue, title=Example Environment Abstraction for the Button Press Task, breakable]
\begin{lstlisting}
class SawyerButtonPressEnvV2(gym.Env):
    def __init__(self):
        self.robot: Robot  # the Sawyer robot in the environment
        self.button: RigidObject  # the button object in the environment
        self.goal_position: np.ndarray[(3,)]  # 3D position of the goal (button pressed position)
        self.trajectory: Trajectory  # stores the trajectory of the episode

    def reset(self) -> np.ndarray:
        # Reset the environment and return initial observation

    def step(self, action: np.ndarray) -> tuple:
        # Perform one step and return (observation, reward, done, info)

    def get_trajectory(self) -> Trajectory:
        # Return the recorded trajectory

class Robot:
    def __init__(self):
        self.ee_position: np.ndarray[(3,)]  # 3D position of the end-effector
        self.joint_positions: np.ndarray[(7,)]  # 7 joint positions of Sawyer robot
        self.joint_velocities: np.ndarray[(7,)]  # 7 joint velocities of Sawyer robot

class RigidObject:
    def __init__(self):
        self.position: np.ndarray[(3,)]  # 3D position of the object (button)
        self.quaternion: np.ndarray[(4,)]  # quaternion of the object (button)

class Trajectory:
    def __init__(self, max_length=25):
        self.states: deque  # queue of states, max length 25
        self.actions: deque  # queue of actions, max length 25
        self.observations: deque  # queue of observations, max length 25

    def add_step(self, state: dict, action: np.ndarray, observation: np.ndarray):
        # Add a step to the trajectory

    def __len__(self) -> int:
        # Return the number of steps in the trajectory

class State:
    def __init__(self):
        self.robot: Robot  # state of the robot
        self.button: RigidObject  # state of the button
\end{lstlisting}
\end{tcolorbox}

\begin{tcolorbox}[colback=lightlightgray, colframe=lightblue, title=Example Environment Abstraction for the Feeding Task, breakable]
\begin{lstlisting}
class FeedingEnv:
    physics: Physics
    task: FeedingTask
    control_timestep: float = 0.02  # 50Hz control frequency
    time_limit: float = 4.0  # 200 steps * 0.02s per step
    
    def step(self, action: np.ndarray):
        """Executes one timestep of the environment with given action and updates trajectory."""
        pass
    
    def reset(self):
        """Resets environment to initial state with new food particles and tool position."""
        pass
        
    def get_trajectory(self) -> Trajectory:
        """Returns trajectory data including states, actions, and observations."""
        pass

class Physics:
    def get_tool_state(self) -> tuple:
        """Returns spoon position and orientation relative to target (mouth)."""
        pass
        
    def get_contact_forces(self) -> tuple:
        """Returns forces applied by robot and tool on human."""
        pass
        
    def get_food_state(self) -> tuple:
        """Returns positions and status of all food particles."""
        pass
        
    def get_end_effector_state(self) -> tuple:
        """Returns end effector position, orientation and velocity."""
        pass
        
    def get_human_state(self) -> tuple:
        """Returns human head position, orientation and joint angles."""
        pass
        
    def get_robot_state(self) -> tuple:
        """Returns robot joint angles and configurations."""
        pass

class FeedingTask:
    trajectory: Trajectory
    target_pos: np.ndarray  # 3D target position (mouth)
    total_food_count: int  # Initial number of food particles
    
    def get_observation(self, physics: Physics) -> collections.OrderedDict:
        """Compiles relevant observations for the feeding task."""
        pass
        
    def get_state(self, physics: Physics) -> dict:
        """Returns complete state information including:
        - Tool position/orientation
        - End effector state
        - Forces on human
        - Food particles state 
   
\end{lstlisting}
\end{tcolorbox}

\section{Example of LLM-based Evaluation Functions} \label{appendix:D}
In this section, we demonstrate example LLM-based evaluation functions during experiments, especially for the Walker and Button Press Tasks. Note that all example evaluation functions were sampled from multiple independent calls from the gpt-4 model. For comparison, we also provide the expert-engineered reward functions for these tasks, which serve as the evaluation basis of the Scripted Teachers baseline. The expert-tuned reward functions and sample functions are shown below.

\begin{tcolorbox}[colback=lightlightgray, colframe=lightblue, title=Expert-engineered Reward Function for the Walker Task, breakable]
\begin{lstlisting}
  def get_reward(self, physics):
    """Returns a reward to the agent."""
    standing = rewards.tolerance(physics.torso_height(),
                                 bounds=(_STAND_HEIGHT, float('inf')),
                                 margin=_STAND_HEIGHT/2)
    upright = (1 + physics.torso_upright()) / 2
    stand_reward = (3*standing + upright) / 4
    if self._move_speed == 0:
      return stand_reward
    else:
      move_reward = rewards.tolerance(physics.horizontal_velocity(),
                                      bounds=(self._move_speed, float('inf')),
                                      margin=self._move_speed/2,
                                      value_at_margin=0.5,
                                      sigmoid='linear')
      return stand_reward * (5*move_reward + 1) / 6
\end{lstlisting}
\end{tcolorbox}

\begin{tcolorbox}[colback=lightlightgray, colframe=lightblue, title=Example LLM-based Evalution Functions for the Walker Task, breakable]
\begin{lstlisting}
####Evalution Function Sample 1####

    def evaluate_trajectory(trajectory: Trajectory) -> float:
        def gaussian(x, mu, sigma):
            return np.exp(-0.5 * ((x - mu) / sigma) ** 2)
    
        def sigmoid(x, k=1):
            return 1 / (1 + np.exp(-k * x))
    
        if len(trajectory.states) == 25:
            a = trajectory.actions[0] is None
            if not a:
                immediate_scores = []
                stability_scores = []
                efficiency_scores = []
                goal_achievement_scores = []
    
                for i, (state, action) in enumerate(zip(trajectory.states, trajectory.actions)):
                    # Immediate evaluation
                    upright_score = (1 + state['torso_upright']) / 2
                    height_score = gaussian(state['torso_height'], mu=trajectory.task._STAND_HEIGHT, sigma=0.1)
                    action_magnitude = np.linalg.norm(action)
                    action_score = gaussian(action_magnitude, mu=0, sigma=0.5)
                    
                    immediate_score = (
                        0.4 * upright_score +
                        0.4 * height_score +
                        0.2 * action_score
                    )
                    immediate_scores.append(immediate_score)
                    
                    # Stability evaluation
                    if i >= 2:
                        height_changes = [trajectory.states[j+1]['torso_height'] - trajectory.states[j]['torso_height'] for j in range(i-2, i)]
                        upright_changes = [trajectory.states[j+1]['torso_upright'] - trajectory.states[j]['torso_upright'] for j in range(i-2, i)]
                        height_stability = 1 - abs(height_changes[1] - height_changes[0])
                        upright_stability = 1 - abs(upright_changes[1] - upright_changes[0])
                        stability_scores.append((height_stability + upright_stability) / 2)
                    
                    # Efficiency evaluation
                    if i >= 1:
                        velocity_change = state['horizontal_velocity'] - trajectory.states[i-1]['horizontal_velocity']
                        efficiency_scores.append(gaussian(velocity_change, mu=0, sigma=0.1))
                    
                    # Goal achievement evaluation
                    target_speed = trajectory.task._move_speed
                    if target_speed == 0:
                        stand_score = gaussian(state['torso_height'], mu=trajectory.task._STAND_HEIGHT, sigma=0.1)
                        upright_score = (1 + state['torso_upright']) / 2
                        goal_achievement_scores.append((stand_score + upright_score) / 2)
                    else:
                        goal_achievement_scores.append(gaussian(state['horizontal_velocity'], mu=target_speed, sigma=target_speed / 4))
                
                # Holistic evaluation
                overall_stability = np.mean(stability_scores)
                overall_efficiency = np.mean(efficiency_scores)
                goal_progression = np.polyfit(range(len(goal_achievement_scores)), goal_achievement_scores, 1)[0]
                goal_progression_score = sigmoid(goal_progression, k=10)
                motion_consistency = 1 - np.std(efficiency_scores)
                task_completion = np.mean(goal_achievement_scores[-10:])
                
                holistic_score = (
                    0.2 * overall_stability +
                    0.2 * overall_efficiency +
                    0.2 * goal_progression_score +
                    0.2 * motion_consistency +
                    0.2 * task_completion
                )
                
                # Combine immediate and holistic scores
                final_score = 0.4 * np.mean(immediate_scores) + 0.6 * holistic_score
            else:
                final_score = 0
        else:
            final_score = 0

        return final_score

        
####Evalution Function Sample 2####

    def evaluate_trajectory(trajectory: Trajectory) -> float:
        """
        Evaluate a given robot trajectory and return an overall score.
        Higher scores indicate better performance.
        
        :param trajectory: The Trajectory object containing states, actions, and observations.
        :return: A single float value representing the overall score.
        """
        # Define weight constants
        WEIGHT_UPRIGHT = 1.0
        WEIGHT_HEIGHT = 0.5
        WEIGHT_VELOCITY = 1.0
        WEIGHT_ENERGY = 0.2
        WEIGHT_STABILITY = 0.8
        WEIGHT_PROGRESS = 0.5
    
        # Immediate evaluation variables
        upright_scores = []
        height_scores = []
        velocity_scores = []
        energy_scores = []
    
        # Holistic evaluation variables
        total_distance = 0
        height_variation = []
        upright_variation = []
    
        target_speed = trajectory.task._move_speed
        stand_height = _STAND_HEIGHT
        max_height_threshold = 0.8 * stand_height  # Threshold below which walker is considered fallen
        
        previous_state = None
        if len(trajectory.states) == 25 :
            a = trajectory.actions[0] == None
            if  type(a) != bool: 
                for step in range(len(trajectory)):
                    state = trajectory.states[step]
                    action = trajectory.actions[step]
                    observation = trajectory.observations[step]
                    
                    # Immediate evaluations
                    torso_upright = state['torso_upright']
                    torso_height = state['torso_height']
                    horizontal_velocity = state['horizontal_velocity']
                    
                    # Upright score: closer to 1 is better
                    upright_scores.append(torso_upright)
                    
                    # Height score: closer to stand height is better
                    height_scores.append(1 - abs(torso_height - stand_height) / stand_height)
                    
                    # Velocity score: closer to target speed is better
                    velocity_scores.append(1 - abs(horizontal_velocity - target_speed) / target_speed)
                    
                    # Energy score: lower action magnitudes are better
                    energy_scores.append(1 - np.linalg.norm(action) / np.sqrt(len(action)))
    
                    # Holistic evaluations
                    if previous_state is not None:
                        distance_travelled = abs(state['horizontal_velocity']) * _CONTROL_TIMESTEP
                        total_distance += distance_travelled
                    
                    height_variation.append(torso_height)
                    upright_variation.append(torso_upright)
                    
                    previous_state = state
    
                # Compute immediate evaluation scores
                mean_upright_score = np.mean(upright_scores)
                mean_height_score = np.mean(height_scores)
                mean_velocity_score = np.mean(velocity_scores)
                mean_energy_score = np.mean(energy_scores)
    
                # Compute holistic evaluation scores
                stability_score = 1 - (np.std(height_variation) / stand_height + np.std(upright_variation)) / 2
                progress_score = total_distance / (len(trajectory) * _CONTROL_TIMESTEP * target_speed)
    
                # Calculate overall score with weights
                final_score = (WEIGHT_UPRIGHT * mean_upright_score +
                                WEIGHT_HEIGHT * mean_height_score +
                                WEIGHT_VELOCITY * mean_velocity_score +
                                WEIGHT_ENERGY * mean_energy_score +
                                WEIGHT_STABILITY * stability_score +
                                WEIGHT_PROGRESS * progress_score)
            else:
                final_score = 0
        else:
            final_score = 0
        return final_score

        
####Evalution Function Sample 3####

    def evaluate_trajectory(trajectory: 'Trajectory') -> float:
        """
        Evaluate the robot's trajectory and return an overall score.
    
        Args:
            trajectory (Trajectory): The trajectory to be evaluated.
    
        Returns:
            float: The overall score of the trajectory.
        """
        
        # Initialize score components
        immediate_scores = []
        total_distance = 0.0
        total_effort = 0.0
        last_velocity = None
        smoothness_penalty = 0.0
        consistency_penalty = 0.0
        
        # Define constants for evaluation thresholds and weights
        TOROS_UPRIGHT_THRESHOLD = 0.8  # Close to fully upright
        TOROS_HEIGHT_THRESHOLD = 1.0  # Close to the stand height
        TARGET_VELOCITY = 1.0  # Target walking speed
        EFFICIENCY_WEIGHT = 0.1  # Weight for efficiency in the overall score
        SMOOTHNESS_WEIGHT = 0.2  # Weight for smoothness in the overall score
        FALL_PENALTY = -100.0  # Penalty for falling
    
        for step in range(len(trajectory)):
            state = trajectory.states[step]
            action = trajectory.actions[step]
            observation = trajectory.observations[step]
    
            # Immediate evaluation
            upright_score = max(state['torso_upright'], 0)  # Prefer upright posture
            height_score = max(0, 1 - abs(state['torso_height'] - trajectory.task._STAND_HEIGHT))
            speed_score = max(0, 1 - abs(state['horizontal_velocity'] - trajectory.task._move_speed))
    
            immediate_score = (upright_score + height_score + speed_score) / 3
            immediate_scores.append(immediate_score)
    
            # Holistic evaluation components
            total_distance += state['horizontal_velocity'] * trajectory.task.control_timestep
            total_effort += np.sum(np.square(action))
    
            if last_velocity is not None:
                smoothness_penalty += np.linalg.norm(state['velocity'] - last_velocity)
            last_velocity = state['velocity']
    
            # Consistency in joint orientations
            consistency_penalty += np.var(observation['orientations'])
    
        # Holistic evaluation
        average_immediate_score = np.mean(immediate_scores)
        efficiency_score = total_distance / (total_effort + 1e-6)  # Avoid division by zero
        smoothness_score = 1 / (smoothness_penalty + 1e-6)  # Smoothness as inverse of penalty
        consistency_score = 1 / (consistency_penalty + 1e-6)  # Consistency as inverse of penalty
    
        final_score = (
            average_immediate_score +
            EFFICIENCY_WEIGHT * efficiency_score +
            SMOOTHNESS_WEIGHT * smoothness_score +
            (1 - EFFICIENCY_WEIGHT - SMOOTHNESS_WEIGHT) * consistency_score
        )
    
        # Penalize for falling
        if trajectory.states[-1]['torso_height'] < 0.8 * trajectory.task._STAND_HEIGHT:
            final_score += FALL_PENALTY
    
        return final_score

        
####Evalution Function Sample 4####

    def evaluate_trajectory(trajectory: Trajectory) -> float:
        """
        Evaluate the robot's trajectory and return an overall score.
    
        Args:
            trajectory (Trajectory): The trajectory to be evaluated.
    
        Returns:
            float: The overall score of the trajectory.
        """
        
        # Constants
        TARGET_HEIGHT = trajectory.task._STAND_HEIGHT  # 1.2 meters
        TARGET_SPEED = trajectory.task._move_speed  # Depends on the current task: 0, 1 m/s, or 8 m/s
        FALL_THRESHOLD = 0.8 * TARGET_HEIGHT  # Around 0.96 meters
        
        # Initialize score components
        stability_score = 0
        speed_score = 0
        efficiency_score = 0
        smoothness_score = 0
        progress_score = 0
        consistency_score = 0
    
        # Initialize counters
        steps = len(trajectory)
        distance_covered = 0
        previous_velocity = None
        previous_orientations = None
    
        # Immediate Evaluation
        for i, (state, action, observation) in enumerate(zip(trajectory.states, trajectory.actions, trajectory.observations)):
            # Stability
            torso_upright = state['torso_upright']
            torso_height = state['torso_height']
            stability_score += max(0, torso_upright) * max(0, (torso_height - FALL_THRESHOLD) / (TARGET_HEIGHT - FALL_THRESHOLD))
    
            # Speed
            horizontal_velocity = state['horizontal_velocity']
            speed_score += max(0, 1 - abs(horizontal_velocity - TARGET_SPEED) / TARGET_SPEED)
    
            # Efficiency
            efficiency_score += 1 - np.linalg.norm(action) / np.sqrt(3)  # Normalized to [0, 1]
    
            # Progress
            if i > 0:
                distance_covered += horizontal_velocity * trajectory.task.control_timestep
    
            # Smoothness (difference in orientations and velocities between consecutive steps)
            if previous_orientations is not None:
                orientation_diff = np.linalg.norm(state['orientations'] - previous_orientations)
                velocity_diff = np.linalg.norm(state['velocity'] - previous_velocity)
                smoothness_score += 1 / (1 + orientation_diff + velocity_diff)
    
            previous_velocity = state['velocity']
            previous_orientations = state['orientations']
    
        # Normalize immediate scores
        if steps > 0:
            stability_score /= steps
            speed_score /= steps
            efficiency_score /= steps
            smoothness_score /= (steps - 1) if steps > 1 else 1
    
        # Holistic Evaluation
        # Progress
        progress_score = distance_covered
    
        # Consistency (variation in orientations and velocities)
        orientation_variation = np.var([state['orientations'] for state in trajectory.states], axis=0).mean()
        velocity_variation = np.var([state['velocity'] for state in trajectory.states], axis=0).mean()
        consistency_score = 1 / (1 + orientation_variation + velocity_variation)
    
        # Combine scores
        final_score = (
            0.3 * stability_score +
            0.3 * speed_score +
            0.1 * efficiency_score +
            0.1 * smoothness_score +
            0.1 * progress_score +
            0.1 * consistency_score
        )
    
        return final_score

\end{lstlisting}
\end{tcolorbox}

\begin{tcolorbox}[colback=lightlightgray, colframe=lightblue, title=Expert-engineered Reward Function for the Button Press Task, breakable]
\begin{lstlisting}
     def compute_reward(
            self, action: npt.NDArray[Any], obs: npt.NDArray[np.float64]
        ) -> tuple[float, float, float, float, float, float]:
            assert (
                self._target_pos is not None
            ), "`reset_model()` must be called before `compute_reward()`."
            del action
            obj = obs[4:7]
            tcp = self.tcp_center
    
            tcp_to_obj = float(np.linalg.norm(obj - tcp))
            tcp_to_obj_init = float(np.linalg.norm(obj - self.init_tcp))
            obj_to_target = abs(self._target_pos[1] - obj[1])
    
            tcp_closed = max(obs[3], 0.0)
            near_button = reward_utils.tolerance(
                tcp_to_obj,
                bounds=(0, 0.05),
                margin=tcp_to_obj_init,
                sigmoid="long_tail",
            )
            button_pressed = reward_utils.tolerance(
                obj_to_target,
                bounds=(0, 0.005),
                margin=self._obj_to_target_init,
                sigmoid="long_tail",
            )
    
            reward = 2 * reward_utils.hamacher_product(tcp_closed, near_button)
            if tcp_to_obj <= 0.05:
                reward += 8 * button_pressed
    
            return (reward, tcp_to_obj, obs[3], obj_to_target, near_button, button_pressed)
\end{lstlisting}
\end{tcolorbox}

\begin{tcolorbox}[colback=lightlightgray, colframe=lightblue, title=Example LLM-based Evalution Functions for the Button Press Task, breakable]
\begin{lstlisting}
####Evalution Function Sample 1####

    def evaluate_trajectory(trajectory: Trajectory) -> float:
        # Weights for different components of the evaluation
        distance_weight = 0.5
        y_alignment_weight = 0.3
        trend_weight = 0.1
        smoothness_weight = 0.1
    
        distance_score = 0.0
        y_alignment_score = 0.0
        trend_score = 0.0
        smoothness_score = 0.0
    
        previous_ee_position = None
        previous_joint_positions = None
        previous_joint_velocities = None
        if len(trajectory.states) == 25 :
            a = trajectory.actions[0] == None
            if  type(a) != bool: 
                for i, (state, action, observation) in enumerate(zip(trajectory.states, trajectory.actions, trajectory.observations)):
                    robot = state.robot
                    button = state.button

                    # Immediate evaluation: Distance to button
                    distance = np.linalg.norm(robot.ee_position - button.position)
                    distance_score += 1.0 / (1.0 + distance)  # Inverse to give higher scores for smaller distances

                    # Immediate evaluation: Y-coordinate alignment
                    y_alignment = 1.0 - abs(robot.ee_position[1] - button.position[1])
                    y_alignment_score += y_alignment

                    # Holistic evaluation: Trend analysis
                    if previous_ee_position is not None:
                        if (robot.ee_position[1] - previous_ee_position[1]) * (button.position[1] - previous_ee_position[1]) > 0:
                            trend_score += 1.0

                    # Holistic evaluation: Movement smoothness
                    if previous_joint_positions is not None and previous_joint_velocities is not None:
                        joint_position_diff = np.linalg.norm(robot.joint_positions - previous_joint_positions)
                        joint_velocity_diff = np.linalg.norm(robot.joint_velocities - previous_joint_velocities)
                        smoothness_score += 1.0 / (1.0 + joint_position_diff + joint_velocity_diff)

                    previous_ee_position = robot.ee_position
                    previous_joint_positions = robot.joint_positions
                    previous_joint_velocities = robot.joint_velocities
    
                # Normalize scores
                num_steps = len(trajectory)
                if num_steps > 0:
                    distance_score /= num_steps
                    y_alignment_score /= num_steps
                    trend_score /= num_steps
                    smoothness_score /= num_steps
    
                # Final score calculation with weights
                final_score = (distance_weight * distance_score +
                                y_alignment_weight * y_alignment_score +
                                trend_weight * trend_score +
                                smoothness_weight * smoothness_score)
            else:
                final_score = 0
        else:
            final_score = 0
        return final_score

        
####Evalution Function Sample 2####

    def evaluate_trajectory(trajectory: Trajectory) -> float:
        immediate_scores = []
        holistic_scores = []
    
        # Immediate evaluation weights
        weight_tcp_to_obj = 0.4
        weight_button_pressed = 0.6
    
        # Holistic evaluation weights
        weight_success_rate = 0.5
        weight_efficiency = 0.3
        weight_stability = 0.2
    
        # Collect immediate scores
        if len(trajectory.states) == 25 :
            a = trajectory.actions[0] == None
            if  type(a) != bool: 
                for state, action, observation in zip(trajectory.states, trajectory.actions, trajectory.observations):
                    tcp_to_obj = np.linalg.norm(observation[4:7] - observation[:3])  # Distance from hand to button
                    button_pressed = state['obj_pos'][1] <= 0.78  # Check if the button is pressed
    
                    immediate_score = (weight_tcp_to_obj * (1 - tcp_to_obj)) + (weight_button_pressed * button_pressed)
                    immediate_scores.append(immediate_score)
    
                # Calculate holistic scores
                total_steps = len(trajectory)
                successful_steps = sum(1 for state in trajectory.states if state['obj_pos'][1] <= 0.78)
                success_rate = successful_steps / total_steps if total_steps > 0 else 0
    
                # Efficiency: Inverse of the number of steps taken to complete the task
                efficiency = 1 / total_steps if total_steps > 0 else 0
    
                # Stability: Variability in the hand's position (lower variability means higher stability)
                hand_positions = np.array([state['hand_pos'] for state in trajectory.states])
                stability = 1 / np.std(hand_positions) if np.std(hand_positions) > 0 else 0
    
                holistic_score = (weight_success_rate * success_rate) + (weight_efficiency * efficiency) + (weight_stability * stability)
    
                # Combine immediate and holistic scores
                final_score = np.mean(immediate_scores) + holistic_score
            else:
                final_score = 0
        else:
            final_score = 0
        return final_score

        
####Evalution Function Sample 3####        

    def evaluate_trajectory(trajectory: Trajectory) -> float:
        # Initializing variables for evaluation
        total_steps = len(trajectory)
        if total_steps == 0:
            return 0.0
        
        proximity_weight = 0.3
        force_weight = 0.2
        smoothness_weight = 0.2
        completion_weight = 0.3
    
        total_proximity_score = 0.0
        total_force_score = 0.0
        total_smoothness_score = 0.0
        task_completed = False
    
        for i in range(total_steps):
            state = trajectory.states[i]
            action = trajectory.actions[i]
            observation = trajectory.observations[i]
    
            # Extract relevant data
            ee_position = state.robot.ee_position
            button_position = state.button.position
            joint_velocities = state.robot.joint_velocities
            goal_position = state.button.position
    
            # Calculate proximity score (closer is better)
            distance_to_button = np.linalg.norm(ee_position - button_position)
            proximity_score = max(0, 1 - distance_to_button)  # Closer to button gives higher score
            total_proximity_score += proximity_score
    
            # Calculate force score (assume we can check the force applied in some way, here we use placeholder)
            force_applied = np.linalg.norm(action)  # Assuming action contains force information
            ideal_force = 1.0  # Placeholder for ideal force value
            force_score = max(0, 1 - abs(force_applied - ideal_force))
            total_force_score += force_score
    
            # Calculate smoothness score (lower joint velocities indicate smoother movement)
            smoothness_score = max(0, 1 - np.mean(np.abs(joint_velocities)))
            total_smoothness_score += smoothness_score
    
            # Check if the task is completed
            if np.array_equal(ee_position, goal_position):
                task_completed = True
    
        # Normalize scores
        avg_proximity_score = total_proximity_score / total_steps
        avg_force_score = total_force_score / total_steps
        avg_smoothness_score = total_smoothness_score / total_steps
        completion_score = 1.0 if task_completed else 0.0
    
        # Calculate overall score
        final_score = (proximity_weight * avg_proximity_score +
                         force_weight * avg_force_score +
                         smoothness_weight * avg_smoothness_score +
                         completion_weight * completion_score)
    
        return final_score

      
####Evalution Function Sample 4####

    def evaluate_trajectory(trajectory: Trajectory) -> float:
        """
        Evaluate the trajectory of the robot based on immediate and holistic criteria.
        
        Parameters:
        trajectory (Trajectory): The trajectory object containing states, actions, and observations.
        
        Returns:
        float: The final score representing the quality of the trajectory.
        """
        # Initialize scores
        proximity_score = 0.0
        action_effectiveness_score = 0.0
        button_press_success_score = 0.0
        consistency_score = 0.0
        efficiency_score = 0.0
        smoothness_score = 0.0
        
        # Immediate evaluation parameters
        proximity_threshold = 0.05  # Threshold for proximity to the button
        button_press_threshold = 0.02  # Threshold for button press success
        
        # Holistic evaluation parameters
        max_steps = 25  # Maximum number of steps in the trajectory
        smoothness_weight = 0.1  # Weight for smoothness in the overall score
        
        # Iterate over the trajectory
        if len(trajectory.states) == 25 :
            a = trajectory.actions[0] == None
            if  type(a) != bool: 
                for i in range(len(trajectory)):
                    state = trajectory.states[i]
                    action = trajectory.actions[i]
                    observation = trajectory.observations[i]
                    
                    # Immediate evaluation
                    tcp_to_obj = np.linalg.norm(state['hand_pos'] - state['obj_pos'])
                    obj_to_target = abs(state['obj_pos'][1] - 0.78)  # Goal y-coordinate is 0.78
                    
                    # Proximity to the button
                    if tcp_to_obj <= proximity_threshold:
                        proximity_score += 1.0
                    
                    # Action effectiveness
                    if i > 0:
                        prev_state = trajectory.states[i-1]
                        prev_tcp_to_obj = np.linalg.norm(prev_state['hand_pos'] - prev_state['obj_pos'])
                        if tcp_to_obj < prev_tcp_to_obj:
                            action_effectiveness_score += 1.0
                    
                    # Button press success
                    if obj_to_target <= button_press_threshold:
                        button_press_success_score += 1.0
                
                # Holistic evaluation
                total_steps = len(trajectory)
                
                # Consistency: How often the robot moves closer to the goal
                for i in range(1, total_steps):
                    prev_state = trajectory.states[i-1]
                    curr_state = trajectory.states[i]
                    prev_tcp_to_obj = np.linalg.norm(prev_state['hand_pos'] - prev_state['obj_pos'])
                    curr_tcp_to_obj = np.linalg.norm(curr_state['hand_pos'] - curr_state['obj_pos'])
                    if curr_tcp_to_obj < prev_tcp_to_obj:
                        consistency_score += 1.0
                
                # Efficiency: Reward quicker task completion
                efficiency_score = max(0, max_steps - total_steps)
                
                # Smoothness: Penalize erratic movements
                for i in range(2, total_steps):
                    prev_action = trajectory.actions[i-1]
                    curr_action = trajectory.actions[i]
                    action_diff = np.linalg.norm(curr_action - prev_action)
                    smoothness_score -= smoothness_weight * action_diff
                
                # Normalize scores
                total_possible_steps = max_steps - 1
                proximity_score /= total_possible_steps
                action_effectiveness_score /= total_possible_steps
                button_press_success_score /= total_possible_steps
                consistency_score /= total_possible_steps
                
                # Combine scores into a final score
                final_score = (
                    proximity_score * 0.3 +
                    action_effectiveness_score * 0.2 +
                    button_press_success_score * 0.3 +
                    consistency_score * 0.1 +
                    efficiency_score * 0.05 +
                    smoothness_score * 0.05
                )
            else:
                final_score = 0
        else:
            final_score = 0
        return final_score    

\end{lstlisting}
\end{tcolorbox}

\begin{tcolorbox}[colback=lightlightgray, colframe=lightblue, title=Example LLM-based Evalution Functions for the Feeding Jaco, breakable]
\begin{lstlisting}
####Evalution Function Sample####

    import numpy as np
    from typing import Dict, List, Tuple
    
    def evaluate_trajectory(trajectory: Dict[str, np.ndarray]) -> float:
        """
        Evaluates a robot feeding trajectory considering task completion, safety, efficiency, and stability.
        
        Args:
            trajectory: Dictionary containing arrays of states and actions over time
                spoon_pos_real: (T, 3) array of spoon positions
                spoon_orient_real: (T, 4) array of spoon orientations (quaternion)
                target_pos_real: (T, 3) array of target (mouth) positions
                robot_joint_angles: (T, 7) array of robot joint angles
                head_pos_real: (T, 3) array of head positions
                head_orient_real: (T, 4) array of head orientations
                spoon_force_on_human: (T, 1) array of forces applied to human
                actions: (T, 7) array of robot actions
                
        Returns:
            float: Final evaluation score (0 to 100)
        """
        # Extract trajectory components
        T = len(trajectory[:3])
        
        # 1. Distance-to-target evaluation (30% of total score)
        distances = np.linalg.norm(trajectory[:3] - trajectory[7:10], axis=1)
        min_distance = np.min(distances)  # Best achieved distance
        final_distance = distances[-1]    # Final distance
        
        # Convert distances to scores (exponential decay)
        distance_score = 30.0 * (0.7 * np.exp(-5.0 * min_distance) + 
                                0.3 * np.exp(-5.0 * final_distance))
        
        # 2. Movement smoothness (20% of total score)
        # Compute velocities and accelerations
        dt = 0.02  # 50Hz control frequency
        velocities = np.diff(trajectory[:3], axis=0) / dt
        accelerations = np.diff(velocities, axis=0) / dt
        
        # Penalize jerk (rate of change of acceleration)
        jerk = np.diff(accelerations, axis=0) / dt
        smoothness_score = 20.0 * np.exp(-0.1 * np.mean(np.linalg.norm(jerk, axis=1)))
        
        # 3. Safety evaluation (25% of total score)
        # Evaluate contact forces
        max_acceptable_force = 5.0  # Newtons
        force_penalties = np.clip(trajectory[24] / max_acceptable_force, 0, 1)
        safety_score = 25.0 * (1.0 - np.mean(force_penalties))
        
        # 4. Spoon orientation stability (15% of total score)
        # Convert quaternions to euler angles for simplicity
        def quat_to_euler(q):
            # Simple quaternion to euler conversion for stability calculation
            # This is a simplified version - you might want to use a proper conversion
            x, y, z, w = q
            roll = np.arctan2(2*(w*x + y*z), 1 - 2*(x*x + y*y))
            pitch = np.arcsin(2*(w*y - z*x))
            yaw = np.arctan2(2*(w*z + x*y), 1 - 2*(y*y + z*z))
            return np.array([roll, pitch, yaw])
        
        spoon_angles = np.array([quat_to_euler(q) for q in trajectory[3:7]])
        angle_changes = np.diff(spoon_angles, axis=0)
        stability_score = 15.0 * np.exp(-2.0 * np.mean(np.abs(angle_changes)))
        
        # 5. Efficiency evaluation (10% of total score)
        # Penalize excessive joint movements
        joint_velocities = np.diff(trajectory[10:17], axis=0) / dt
        energy_expenditure = np.mean(np.sum(joint_velocities**2, axis=1))
        efficiency_score = 10.0 * np.exp(-0.1 * energy_expenditure)
        
        # Combine all scores
        final_score = (distance_score + smoothness_score + safety_score + 
                      stability_score + efficiency_score)
        
        # Normalize to 0-100 range
        final_score = np.clip(final_score, 0, 100)
        
        return float(final_score)


\end{lstlisting}
\end{tcolorbox}

We can observe that compared to the expert-designed reward functions, the LLM-based evaluation functions cover more than just success-related criteria, providing a more nuanced evaluation pattern. Also, as required by the prompts, the LLM-based evaluation functions cover immediate state-action pairs as well as holistic evaluations.

For example, on the Walker task, the expert reward function is primarily focused on immediate task success, measured through Upright Posture, which rewards the walker for keeping its torso upright, and Torso Height, which ensures the torso height is within a certain range. On the other hand, the LLM-based evaluation function integrates these success-related criteria but also extends the evaluation to cover additional aspects, such as Energy Efficiency, measured by penalizing large action magnitudes to promote energy-efficient behavior, and Stability Over Time by evaluating changes in torso height and uprightness, ensuring stability throughout the trajectory. By incorporating broader criteria—both immediate and holistic—the LLM-based evaluation functions provide a more comprehensive and nuanced assessment of the robot trajectories. This ensures the walker not only completes the tasks successfully but also does so efficiently, stably, and consistently over time, leading to potentially more robust and effective reinforcement learning outcomes.

More importantly, we observe that the evaluation functions generated from the same gpt-4o agents, exhibit diversity. This variation manifests in several ways, such as differing task-related criteria, assorted definitions for the same criteria, and varying priorities assigned to these criteria (e.g., different weighting schemes). Our PrefCLM capitalizes on this diversity, leveraging the unique understanding that each LLM agent brings to the task and leading to a richer and more comprehensive evaluation process.

\section{The Feeding Task in the User Study} \label{appendix:E}

We selected the Feeding task from the Assistive Gym: A Physics Simulation Framework for Assistive Robotics \cite{erickson2020assistive}. In this task, a robot arm is tasked with delivering a spoon holding food, represented as small spheres, to the mouth of a human seated in a chair without spilling.
\subsection{Pre-trained Robot Policy}
To pre-train a robot policy, we utilized the ground-truth reward functions provided by the benchmark, which consist of several costs and penalties to differentiate:
\begin{itemize}
    \item \( C_d(s) \): cost for long distance from the robot’s end effector to the target assistance location (e.g., human mouth).
    \item \( C_e(s) \): reward for successfully feeding food to the human mouth.
    \item \( C_v(s) \): cost for high robot end effector velocities.
    \item \( C_f(s) \): cost for applying force away from the target assistance location.
    \item \( C_{hf}(s) \): cost for applying high forces near the target.
    \item \( C_{fd}(s) \): cost for spilling food on the human.
    \item \( C_{fdv}(s) \): cost for food entering the mouth at high velocities.
\end{itemize}

We selected the default weights for these criteria as in \cite{erickson2020assistive}. We trained the robot policy using Soft Actor-Critic (SAC) for a total of \(1.6 \times 10^7\) time steps, approximately 8 hours. SAC is also the RL training basis of PEBBLE \cite{lee2021pebble}, the PbRL backbone algorithm for our PrefCLM, ensuring smooth fine-tuning with PrefCLM. After pre-training, the robot policy is capable of basic functionality, i.e., successfully bringing the spoon close to a certain distance from the user’s mouth.

Furthermore, the Assistive Gym allows for adjusting the human shape, location of the chair, mounting of the robotic arm, and other physical parameters, providing a good opportunity to mimic realistic settings during pre-training.

\subsection{Fine-tuning Process}
During the user study, we aimed to fine-tune the pre-trained robot policy using PrefCLM (few-shot, n=10) by incorporating user interactive feedback and compare the resulting satisfaction and personalization against the baseline PrefEVO and pre-trained robot policy.

Each participant first expressed their initial expectations for the Feeding Task in natural language, e.g., ``I want the robot to move carefully and slowly when feeding me." PrefCLM then generated initial evaluation functions based on these expectations. Using the crowdsourced evaluation functions, we fine-tuned the pre-trained policy.

For each model (PrefCLM and baseline PrefEVO), we periodically (every $4 \times 10^6$ environment steps, approximately 2 hours) rolled out the learned robot policy to the physical Kinova Jaco robotic arm, for a total of three times. Each time, participants provided interactive feedback, which was utilized to refine the evaluation functions.

Specifically, we conducted the following steps:

\begin{itemize}
\item Fine-tuned the pre-trained policy for $4 \times 10^6$ environment steps using the initial evaluation functions generated based on user expectations.
\item Participants interacted with the first fine-tuned policy and provided the first interactive feedback.
\item PrefCLM adapted the evaluation functions based on this feedback and fine-tuned the policy again for $4 \times 10^6$ environment steps.
\item Repeated the interaction and feedback process with the second fine-tuned policy.
\item PrefCLM adapted the evaluation functions once more and fine-tuned the policy again for $4 \times 10^6$ environment steps.
\item Repeated the interaction and feedback process with the third fine-tuned policy.
\item PrefCLM adapted the evaluation functions once more and conducted a final fine-tuning for $4 \times 10^6$ environment steps.
\end{itemize}

It is worth noting that PEBBLE, the PbRL backbone algorithm for our PrefCLM, is an off-policy PbRL algorithm. This means that when the evaluation function is adapted and a new reward model is learned in the PbRL setting, PEBBLE can re-label all state-action pairs from the behavior of previous robot policies and reward models in the replay buffer. This ensures efficient use of previous experiences and accelerate the learning process.

For participants, they could choose to leave or stay during the fine-tuning process. If they decided to leave, we stored the fine-tuned robot policy after current round of training, and resumed with the next round of interaction, interactive feedback, evaluation function adaptations, and fine-tuning upon their return.

The final fine-tuned robot policy by PrefCLM is compared to the final fine-tuned one by PrefEVO and the pre-trained policy, as the final interaction policy. Each participant interacted with each policy three times in a randomized sequence and was not informed about which policy was active to prevent any bias in their responses. Following each interaction, participants were asked to rate the robot behaviors of the three policies in terms of satisfaction, using a Likert scale ranging from 1 (strongly disagree) to 5 (strongly agree). They were also asked to rate the level of personalization resulting from PrefCLM and PrefEVO using the same scale.

\section{\textcolor{black}{Additional Experimental Results}} \label{appendix:F}

\subsection{\textcolor{black}{Numerical Results of the Ablation Study on Agent Configurations (Fig. 5).}}

\begin{table}[th]
   \centering
   \caption{\textcolor{black}{Performance comparison of different methods on button pressing and walker walking tasks. HO: Homogeneous agents, HE: Heterogeneous agents. Values show success rate (\%) for button press and episode returns for walker walk, averaged over 5 runs.}}
   \label{tab:method_comparison}
   \begin{tabular}{lccccccccccccc}
       \toprule
       & \multicolumn{6}{c}{Button Press (Success Rate \%)} & \multicolumn{6}{c}{Walker Walk (Episode Return)} \\
       \cmidrule(lr){2-7} \cmidrule(lr){8-13}
       & \multicolumn{2}{c}{Maximum} & \multicolumn{2}{c}{Mean} & \multicolumn{2}{c}{Final} & \multicolumn{2}{c}{Maximum} & \multicolumn{2}{c}{Mean} & \multicolumn{2}{c}{Final} \\
       Method & Val & Std & Val & Std & Val & Std & Val & Std & Val & Std & Val & Std \\
       \midrule
       Single LLM & 48.00 & 24.00 & 26.78 & 11.72 & 38.40 & 20.00 & 496.11 & 25.50 & 395.56 & 32.60 & 478.62 & 27.05 \\
       PrefCLM, HO, N=3 & 69.99 & 11.68 & 45.67 & 12.21 & 37.33 & 13.35 & 718.95 & 41.80 & 560.21 & 34.34 & 718.95 & 41.80 \\
       PrefCLM, HO, N=10 & 82.50 & 0.00 & 77.54 & 3.68 & 74.80 & 9.10 & 853.39 & 15.18 & 769.51 & 14.03 & 846.78 & 11.31 \\
       PrefCLM, HO, N=20 & 99.48 & 2.48 & 69.95 & 5.22 & 64.03 & 0.29 & 967.38 & 24.50 & 835.63 & 79.74 & 844.80 & 111.00 \\
       MajCLM, HO, N=3 & 69.23 & 17.75 & 32.97 & 8.47 & 31.20 & 8.00 & 615.03 & 78.45 & 483.07 & 65.76 & 597.82 & 71.12 \\
       MajCLM, HO, N=10 & 77.81 & 2.26 & 63.55 & 3.50 & 57.93 & 1.58 & 796.61 & 13.85 & 665.98 & 42.45 & 760.18 & 36.97 \\
       MajCLM, HO, N=20 & 81.25 & 6.50 & 48.07 & 3.81 & 50.00 & 3.80 & 819.74 & 31.62 & 729.76 & 46.28 & 816.04 & 13.58 \\
       PrefCLM, HE, N=3 & 78.14 & 7.57 & 48.75 & 2.14 & 41.24 & 2.11 & 674.39 & 11.89 & 604.70 & 34.09 & 642.12 & 56.78 \\
       MajCLM, HE, N=3 & 51.90 & 6.50 & 43.63 & 7.76 & 37.92 & 8.80 & 599.35 & 16.52 & 536.44 & 26.71 & 569.00 & 18.82 \\
       \bottomrule
   \end{tabular}
   \vspace{-3mm}
\end{table}
\subsection{\textcolor{black}{Experiments on General RL Tasks with HITL Module}}
\textcolor{black}{We further conducted experiments on general RL tasks using two configurations: PrefCLM (few-shot) + HITL and PrefEVO + HITL. In the PrefEVO + HITL setup, we replaced the LLM self-reflection mechanism in PrefEVO's evolutionary search with human feedback, aligning this model with the method described in \cite{xietext2reward}. For both configurations, two rounds of human feedback were provided.}

\begin{figure*}[ht]
\centering
\includegraphics[width=\linewidth]{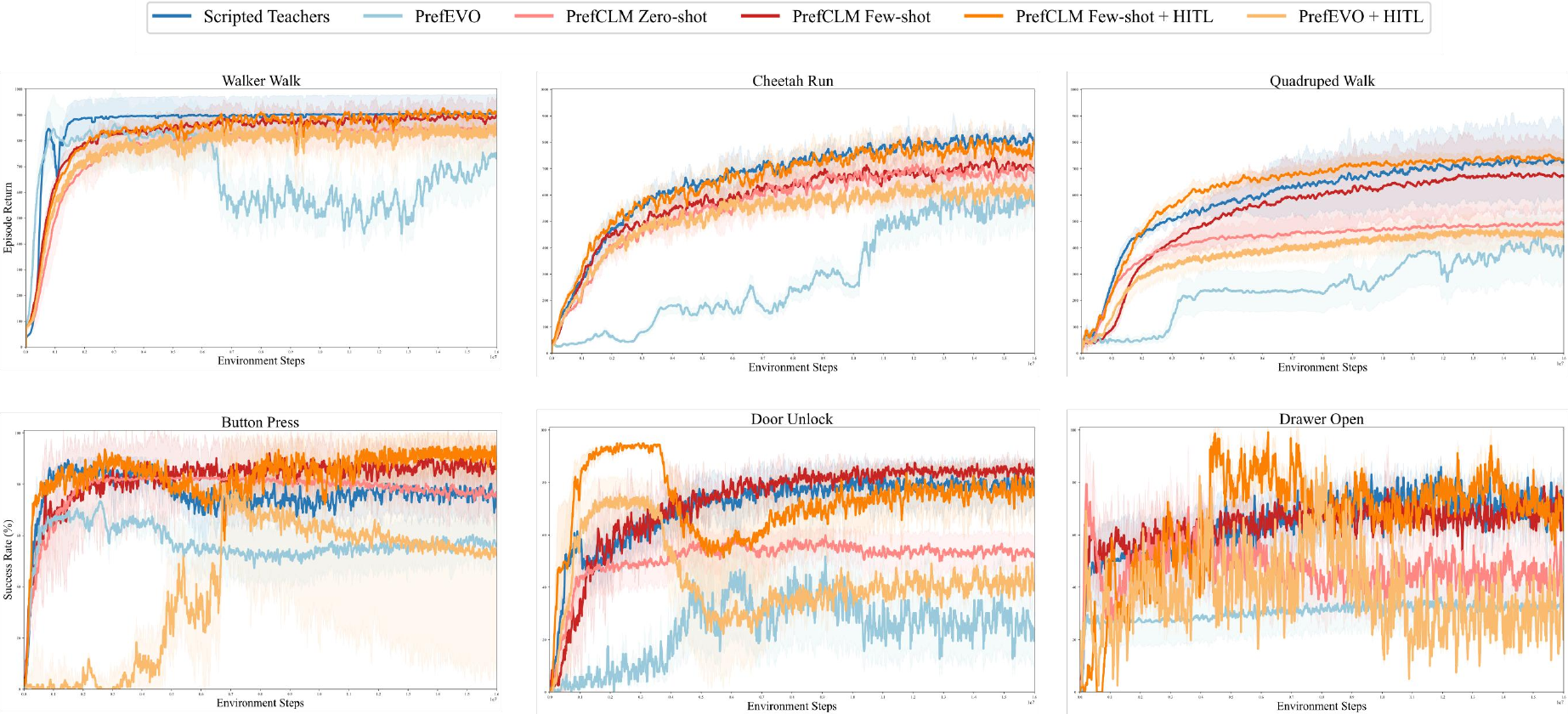}
\caption{\textcolor{black}{Learning curves on general RL tasks, measured in episode returns for locomotion tasks and success rates for manipulation tasks. The solid line represents the mean, while the shaded area indicates the standard deviation across five runs.}}
\label{fig:extra}
\end{figure*}

\textcolor{black}{The results, shown in Fig. \ref{fig:extra}, demonstrate that both PrefCLM and PrefEVO benefit from human-in-the-loop feedback, showing improved performance. However, PrefCLM + HITL consistently outperforms PrefEVO + HITL, highlighting the effectiveness of the crowdsourcing and DST fusion strategies in PrefCLM. These strategies not only enhance task performance but also more effectively capture and leverage human feedback, aligning well with findings from our HRI experiments.}

\subsection{\textcolor{black}{Ablation Study on the Backbone PbRL Algorithm}}

\textcolor{black}{We conducted an ablation study to evaluate the compatibility and performance of PrefCLM (zero-shot, n=10) when integrated with different backbone PbRL algorithms. Specifically, we integrated PrefCLM with two contemporary state-of-the-art PbRL backbones: SURF \cite{park2021surf} and MRN \cite{liu2022meta} using the Drawer Open and Cheetah Run tasks. As shown in Table \ref{backbone}, our findings reveal three key insights: First, PrefCLM seamlessly integrates with different PbRL methods without introducing significant performance overhead. Second, the performance improvements scale proportionally with the capabilities of the underlying PbRL methods, following the established hierarchy (MRN\textgreater SURF\textgreater PEBBLE) demonstrated in \cite{liu2022meta}. Third, PrefCLM maintains consistent effectiveness across different backbone architectures. These results demonstrate PrefCLM's versatility as a plug-and-play enhancement that can robustly augment various PbRL algorithms. }

\begin{table}[h]
\centering
\caption{\textcolor{black}{Performance comparison of PrefCLM with different PbRL backbones on two tasks. For Drawer Open task, values represent success rate (\%); for Cheetah Run task, values represent episode returns. Results are averaged over 5 runs.}}
\label{tab:backbone_ablation}

\begin{tabular}{@{}lcccccc@{}}
\toprule
\multirow{2}{*}{\textbf{Method}} & \multicolumn{3}{c}{\textbf{Drawer Open (Success Rate \%)}} & \multicolumn{3}{c}{\textbf{Cheetah Run (Episode Return)}} \\
\cmidrule(lr){2-4} \cmidrule(lr){5-7}
& Max & Mean & Last & Max & Mean & Last \\
\midrule
PEBBLE & 87.3 ± 15.2 & 51.4 ± 14.0 & 40.3 ± 11.1 & 716.9 ± 43.1 & 565.1 ± 28.4 & 693.2 ± 39.7 \\
SURF & 87.8 ± 18.5 & 49.4 ± 22.5 & 49.5 ± 24.0 & 847.9 ± 70.2 & 645.2 ± 40.7 & 835.9 ± 74.1 \\
MRN & 87.0 ± 14.5 & 46.0 ± 6.0 & 60.0 ± 10.0 & 860.9 ± 80.8 & 686.7 ± 63.3 & 817.6 ± 85.8 \\
\bottomrule
\end{tabular}

\label{backbone}
\end{table}

\subsection{\textcolor{black}{Ablation Study on the Crowd Composition}}

\textcolor{black}{To provide additional insights into how the composition and quality of LLM agents in the crowd influence performance, we conducted an ablation study on the Walker Walk task using PrefCLM-Zero-shot. We tested various combinations of GPT-4 and GPT-3.5 models, assuming GPT-3.5 as the weaker model.}

\textcolor{black}{The results, presented in Table \ref{Comb}, highlight clear performance differences between high-quality and lower-quality agents, as evidenced by the gap in final returns between 3 × GPT-4 (718.95) and 3 × GPT-3.5 (582.18). This confirms that the quality of LLM agents in the crowd does impact overall performance.}

\textcolor{black}{However, our results also demonstrate the effectiveness of our crowdsourcing approach in improving performance, even with weaker models. For example, 3 × GPT-3.5 significantly outperforms 1 × GPT-3.5 in final returns (582.18 vs. 321.66). This indicates that PrefCLM effectively aggregates multiple perspectives, leveraging the strengths of even weaker models.}

\textcolor{black}{Interestingly, while performance decreases as weaker agents are introduced (e.g., transitioning from 3 × GPT-4 to 1 × GPT-4 + 2 × GPT-3.5), our DST fusion strategy exhibits good robustness. For instance, replacing one GPT-4 with GPT-3.5 (2 × GPT-4 + 1 × GPT-3.5) results in nearly identical final returns compared to 3 × GPT-4 (715.31 vs. 718.95). This highlights DST's ability to mitigate the impact of weaker agents when stronger agents remain in the majority.}

\textcolor{black}{Moreover, replacing one weaker agent with a stronger one, as in the case of 1 × GPT-4 + 2 × GPT-3.5 compared to 3 × GPT-3.5, leads to a notable improvement in final returns (+43.64) and a significant reduction in variance (standard deviation: 98.49 vs. 29.93). This demonstrates that DST fusion effectively capitalizes on the strengths of higher-quality agents, even when stronger agents are not in the majority, showcasing the robustness and adaptability of our approach in diverse crowdsourcing scenarios.}

\textcolor{black}{To sum up, our PrefCLM is designed to enhance performance regardless of crowd composition, and our DST fusion strategy effectively handles both heterogeneous agent quality and inter-agent conflicts. The ablation results demonstrate that our method can maintain robust performance even with mixed-quality agents, while effectively leveraging stronger agents when available.}

\begin{table}[h]
   \centering
   \caption{\small \textcolor{black}{Performance comparison of different LLM combinations on the Walker-Walk task. We report the episode returns averaged over 5 runs (\textit{Value}) with standard deviations (\textit{Std}).}}
   \label{tab:llm_combinations}
   \begin{tabular}{lccccccc}
       \toprule
       \multirow{2}{*}{LLM Combination} & \multicolumn{2}{c}{Maximum Return$^1$} & \multicolumn{2}{c}{Mean Return$^2$} & \multicolumn{2}{c}{Final Return$^3$} \\
       \cmidrule(lr){2-3} \cmidrule(lr){4-5} \cmidrule(lr){6-7}
       & Value & Std & Value & Std & Value & Std \\
       \midrule
       3 × GPT-4 & 720.65 & 39.62 & 560.21 & 34.34 & 718.95 & 41.80 \\
       2 × GPT-4 + 1 × GPT-3.5 & 706.31 & 47.78 & 551.69 & 45.39 & 715.31 & 47.78 \\
       1 × GPT-4 + 2 × GPT-3.5 & 664.06 & 48.08 & 588.52 & 44.77 & 625.82 & 29.93 \\
       3 × GPT-3.5 & 651.30 & 43.10 & 570.54 & 90.32 & 582.18 & 98.49 \\
       1 × GPT-3.5 & 382.87 & 63.88 & 377.67 & 36.61 & 321.66 & 56.11 \\
       \bottomrule
       \multicolumn{7}{l}{\footnotesize $^1$Maximum: Best episode return achieved during training} \\
       \multicolumn{7}{l}{\footnotesize $^2$Mean: Average episode return across all training episodes} \\
       \multicolumn{7}{l}{\footnotesize $^3$Final: Episode return after training completion} \\
   \end{tabular}
   \label{Comb}
   \vspace{-3mm}
\end{table}

\end{document}